\documentclass[10pt,twocolumn,letterpaper]{article}
\pdfoutput=1 
\usepackage{iccv}
\usepackage{times}
\usepackage{epsfig}
\usepackage{graphicx}
\usepackage[accsupp]{axessibility} 
\usepackage{tabularx,ragged2e}
\usepackage{amsmath}
\usepackage{multirow}
\usepackage{amssymb}
\usepackage{lipsum}
\usepackage{booktabs}
\usepackage{subfigure}
\usepackage{caption}
\newcommand{\xk}[1]{{#1}}

\newcommand{\zhpcui}[1]{{#1}} 

\newcommand\blfootnote[1]{%
  \begingroup
  \renewcommand\thefootnote{}\footnote{#1}%
  \addtocounter{footnote}{-1}%
  \endgroup
}

\usepackage[breaklinks=true,bookmarks=false]{hyperref}

\iccvfinalcopy 

\def\iccvPaperID{3738} 

\ificcvfinal\fi

\begin{document}

\title{Deep Hybrid Self-Prior for Full 3D Mesh Generation}

\author{Xingkui Wei$^{1*}$ \quad Zhengqing Chen$^{1*}$ \quad Yanwei Fu$^{1\dag}$  \quad Zhaopeng Cui$^{2}$ \quad Yinda Zhang$^{3\dag}$\\
$^{1}$ Fudan University \quad $^{2}$ Zhejiang University \quad $^{3}$ Google
}

\twocolumn[{%
\renewcommand\twocolumn[1][]{#1}%
\maketitle

\begin{center}
    \centering
    \includegraphics[trim=30 30 30 20, width=0.96\textwidth, clip=true]{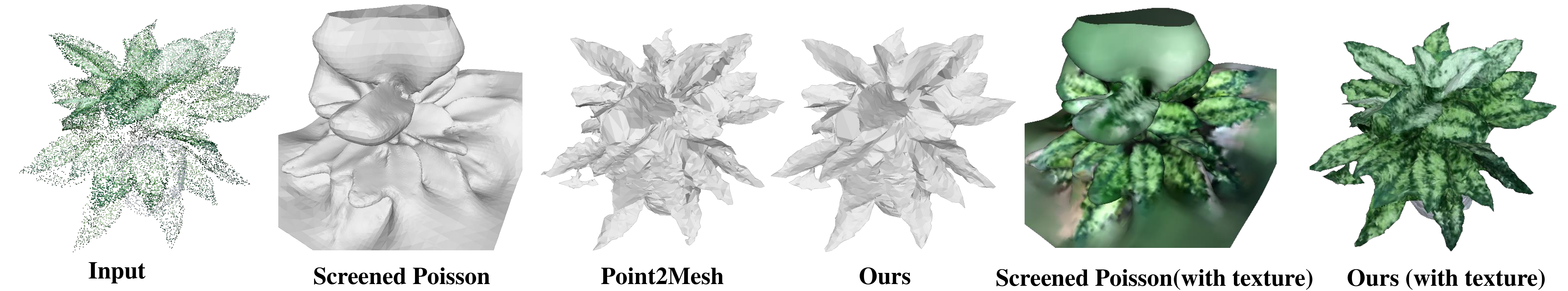}
        \begin{tabularx}{0.95\linewidth}{c *6{>{\Centering}X}}
    \hspace*{0.5cm}Input &\hspace*{1.2cm} Poisson & \hspace*{1.2cm} Point2Mesh & \hspace*{0.3cm} Ours & Poisson &Ours 
    \end{tabularx}
    \vspace{-4mm}
    \captionof{figure}{We propose to utilize the deep hybrid 2D-3D self-prior in neural networks to generate the high-quality textured 3D mesh model from the sparse colored point cloud.}
\end{center}%
}]
\ificcvfinal\thispagestyle{empty}\fi
\begin{abstract}
We present a deep learning pipeline that leverages network self-prior to recover a full 3D model consisting of both a triangular mesh and a texture map from the colored 3D point cloud. 
Different from previous methods either exploiting 2D self-prior for image editing or 3D self-prior for pure surface reconstruction, 
we propose to exploit a novel hybrid 2D-3D self-prior in deep neural networks to significantly improve the geometry quality and produce a high-resolution texture map, which is typically missing from the output of commodity-level 3D scanners.
In particular, we first generate an initial mesh using a 3D convolutional neural network with 3D self-prior, and then encode both 3D information and color information in the 2D UV atlas, which is further refined by 2D convolutional neural networks with the self-prior. In this way, both 2D and 3D self-priors are utilized for the mesh and texture recovery.
Experiments show that, without the need of any additional training data, our method recovers the 3D textured mesh model of high quality from sparse input, and outperforms the state-of-the-art methods in terms of both the geometry and texture quality.
\blfootnote{$^{*}$Equal contribution}
\blfootnote{$^{\dag}$Corresponding author}
\blfootnote{\,\,\,Project page: \url{https://yqdch.github.io/DHSP3D}}
\end{abstract}

\section{Introduction}
Textured mesh is one of the most desirable representation for 3D objects, which has been widely used in many applications, such as industrial design and digital entertainment because it enables not only the 3D related task like collision detection but also the rendering capability.
As a result, the ability to create a full 3D model \zhpcui{consisting of} both a 3D triangular mesh and a texture map is a long-lasting problem and \zhpcui{consistently draws attention.}
\zhpcui{
While purely image-based solution exists~\cite{MultiviewGeometryBook}, the 3D scanners with active illumination usually provide much more accurate 3D models and are robust against challenging cases, e.g. texture-less regions.
Unfortunately, many commodity-level 3D scanners, e.g. Artec Eva~\cite{artec3d}, iReal 2S~\cite{ireal2s}, Einscan Pro~\cite{einscan} etc, only produce colored point clouds as the output, where the object surfaces and texture maps are  missing.
There are plenty of works that generate a mesh model from a point cloud, but they usually rely on strong assumptions~\cite{park2019deepsdf,liu2020meshing}, require pre-training on large dataset~\cite{badki2020meshlet,mi2020ssrnet,erler2020points2surf,chabra2020deep}, and do not produce a texture. 
}

In this work, we propose a method for reconstructing a full 3D model, i.e., a textured triangular mesh, from a colored point cloud.
This task is highly under-constrained, and thus prior knowledge is extremely important.
It is well known that deep learning model is good at learning prior from a large dataset~\cite{park2019deepsdf,erler2020points2surf,chabra2020deep}, but also at the same time prone to overfitting to the dataset bias.
Instead, Ulyanov \etal \cite{ulyanov2018deep} proposed to randomly initialize a convolutional neural network (CNN) to upsample a given image, which used the network structure as a prior without the need of any additional training data.
Sharing the similar spirit, we resort to such self-prior naturally encoded in the neural network for the full 3D model reconstruction task.
As one of the most related prior arts, Hanocka \etal \cite{hanocka2020point2mesh} proposed to create a mesh, without texture, from a point cloud using a MeshCNN~\cite{hanocka2019meshcnn} to deform from the convex hull.
They found that the graph-based CNN can also learn the self-prior from the input point cloud to reconstruct a 3D mesh with noise suppressed and missing parts filled.
Despite significant improvements over previous methods and the capability to handle challenging cases, however, its output quality highly depends on the input noise level and sparsity (See Sec.\ref{sec:surface}), and the effect of 3D network self-prior is not phenomenal as its 2D counterpart \cite{ulyanov2018deep} empirically.

We propose to exploit the hybrid 2D-3D self-prior for full mesh reconstruction. 
Specifically, we first utilize the 3D MeshCNN \cite{hanocka2019meshcnn} to exploit the 3D prior in a similar way as Hanocka \etal \cite{hanocka2020point2mesh} and generate an initial 3D mesh model.
Then we create a UV atlas encoding the 3D location of the points instead of color information, which is then refined by a 2D CNN using the self-prior and used to update the 3D mesh.
We find this 2D network is surprisingly more effective, compared to the 3D MeshCNN, in learning self-prior, and can provide valuable regularization in producing high-resolution mesh with delicate details.
The 3D-prior and 2D-prior network runs iteratively to refine the 3D mesh model, and extensive experiments show that our model significantly improves the geometry quality.

Besides the triangulated mesh, our method also recovers a high-resolution texture map.
While it is not trivial to build such a texture map from colors on the sparse point cloud since the texture maps are usually in much higher resolution than the 3D geometry, e.g., the number of faces, we borrow the help from the 2D self-learned CNN with the belief that the self-prior is stronger and easier to learn on 2D CNN compared to a 3D graph-based convolutional neural network (GCN).
Using the same UV atlas generated from the 3D mesh, we train a 2D CNN to recover the color from sparse point, and find appealing texture maps can be generated automatically. The texture map will be iteratively optimized together with the 3D mesh model.  

Our contributions can be summarized as follows. First, we propose a deep learning pipeline that reconstructs a full 3D model with both a triangular mesh and a texture map from a sparse colored point cloud by leveraging self-prior from the network. Second, a novel hybrid 2D-3D self-prior is exploited in our pipeline without learning on any extra data for both geometry and texture recovery. Experiments demonstrate that our method outperforms both the traditional and the existing state-of-the-art deep learning based methods, and both 2D prior and 3D prior benefits the full mesh reconstruction.

\section{Related work}

\paragraph{Traditional Surface Reconstruction methods}
There is a long history of reconstructing surfaces from point clouds. Early methods such as Delaunay triangulations \cite{boissonnat1984geometric} and Voronoi diagrams~\cite{amenta1998new}
interpolate points by creating a triangular mesh. When there are noises, however, the resulting surface is often jagged. As a result, special data pre-processing is usually required to generate a smooth surface.

Mainstream approaches to reconstruct surface are based on implicit function approaches which can be classified into global and local approaches. Global approaches, such as radial basis functions (RBFs)~\cite{carr2001reconstruction}, consider all the data at once, and define a scalar function which used for testing if a point is inside or outside the surface. In contrast, local approaches, such as truncated signed distance function (TSDF) \cite{curless1996volumetric} and moving least squares (MLS)~\cite{amenta2004defining,levin2004mesh}, consider only subsets of nearby points.

The algorithm of (Screened) Poisson surface reconstruction \cite{kazhdan2006poisson,kazhdan2013screened} combines the advantages of global and local approaches. It finds an indicator function and uses its gradient field to solve the Poisson equation, and then an isosurface can be extracted to reconstruct the surface. The reconstructed model is watertight closed and has good surface details, however this approach requires accurate normal orientation, relies on the dense point cloud, and struggles to handle non-watertight cases.

\paragraph{Deep Learning for Full Model Reconstruction}
Deep learning provides new opportunities for geometric reconstruction, especially in terms of 3D representation.
3D volumes \cite{ChoyXGCS16,GirdharFRG16} and point clouds \cite{fan2017point,achlioptas2017representation,achlioptas2018learning} have been prevalent in many early works, which unfortunately are usually restricted to the low resolutions due to the memory constraint.
Recently, the implicit representations \cite{Occupancy_Networks, park2019deepsdf,jiang2020local,chibane2020implicit,erler2020points2surf,chabra2020deep} have been investigated and greatly improve the geometry details, but demand comparatively long inference time due to the sampling {and suffer from overfitting on training sets. Liu. \etal \cite{liu2020meshing} builds a network to estimate the local connectivity of input points for surface reconstruction, 
whose performance heavily relies on the quality of inputs.}
Similar to us, a graph-based CNN is proposed to generate a 3D model directly in a triangulated mesh that is ready to use. 
A common approach is to deform gradually from an initial shape toward the desired output \cite{hanocka2019meshcnn,pixel2mesh,pixel2mesh++,KatoUH2018,PontesKSLF2017}. 
with the help of learned guidance.

On the other hand, there has been a lot of efforts~\cite{saito2017photorealistic,kanazawa2018learning, zhu2018visual,oechsle2019texture,sun2018im2avatar, huang2020adversarial} on texture generation from 2D images. Nevertheless, very few works focus on learning a texture consistent with the sparse color observed from point clouds, which is still a challenging problem.
\begin{figure*}[t]
\centering \includegraphics[width=0.95\linewidth]{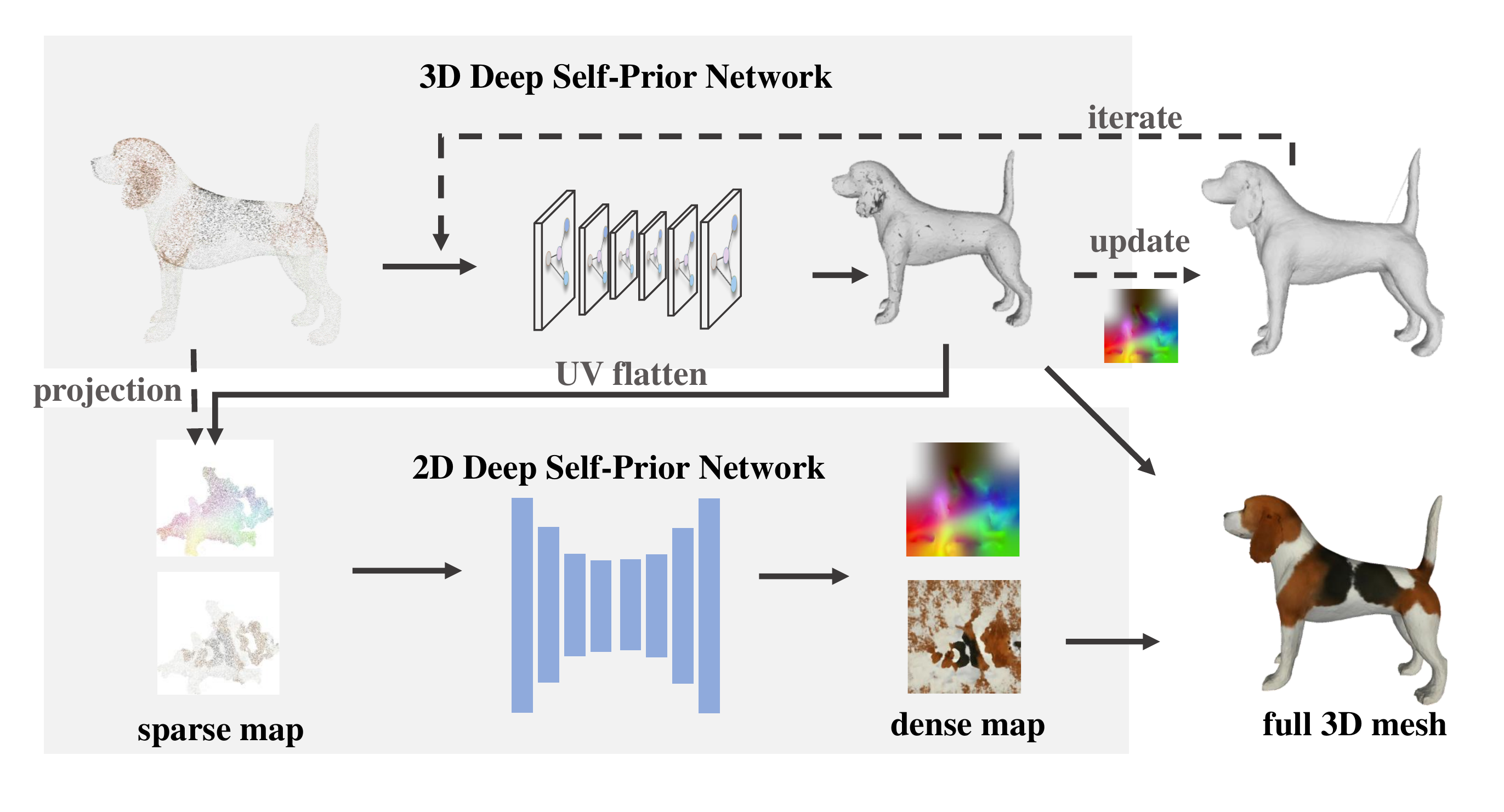}
\vspace{-4mm}
\caption{Overview of our model. Our full model contains two building blocks, namely, 3D deep self-prior network, and 2D deep self-prior network, which run iteratively to improve the geometry and texture outputs.}
\label{fig:overview} 
\vspace{-4mm}
\end{figure*}

\paragraph{Deep Network as a prior}
Recently, the \emph{deep image prior }(DIP)~\cite{ulyanov2018deep} has shown its strong ability for self-supervised 2D image reconstruction tasks (eg. image super-resolution, denoising, or inpainting).
Follow up works \cite{gandelsman2019double,quan2020self2self,ren2020neural} shows the capability of deep prior to capture image statistics for multiple low-level 2D vision tasks such as dehazing, transparency separation, deblur, etc.
Inspired by \cite{ulyanov2018deep}, our approach aims to transfer 2D deep priors to 3D geometry space, which enhances the smoothness and robustness of 3D mesh and texture generation. 

For 3D surface reconstruction, deep geometric prior is presented in \cite{williams2019deep}, which fits MLPs for different local regions of the point clouds. Point2Mesh~\cite{hanocka2020point2mesh} reconstructs a surface mesh from an input point cloud by optimizing a CNN-based self-prior deep network \cite{hanocka2019meshcnn}.
While encoding mesh shape into network parameters, those methods leverage the representational power of the deep network to remove noise. However, the 3D priors of the network from unstructured 3D data is not strong enough. Consequently, the performance is highly relative to the quality of the input point cloud.


\section{Full Mesh Generation with Hybrid Prior}
Given a colored input point cloud, our goal is to reconstruct the corresponding surface mesh with fine details of both the geometry shape and texture. 
We design a hybrid-prior network to leverage self-prior in both 3D and 2D space, and an overview of our system is illustrated in Fig. \ref{fig:overview}.
We first run a MeshCNN based 3D self-prior network to reconstruct an initial untextured 3D mesh, then improve the geometry quality and produce a high-resolution texture map in an iterative fashion.
Particularly in each iteration, we first build a texture atlas using the current geometry mesh, and then warp the location and color of the input sparse point cloud into texture UV space, which generates a sparse location UV map and a sparse color UV map respectively.
\xk{Two separate
2D self-prior networks are trained for XYZ refinement and RGB generation respectively using the sparse maps as supervision.}
We use the predicted dense location UV map to update the vertex location in the 3D mesh, which is then fed into the 3D self-prior network for another refinement.
This hybrid-prior network runs iteratively until the geometry mesh and texture output is stable.

\subsection{3D Prior Network}
\label{sec:3DPrior}
Our model starts from a MeshCNN based 3D-Prior network to create an initial mesh from the input point cloud.
The main purpose is to build a surface manifold which facilitates the use of 2D prior (Sec. \ref{sec:2dprior}).

Similar to Point2Mesh \cite{hanocka2020point2mesh}, given an initial point cloud, a convex hull is generated as an initial mesh which will be deformed to the target shape.
A graph is built on the edges of the initial convex hull, on which a MeshCNN can be run to produce the displacement of vertices that updates the 3D shape.
The feature on each graph node, which corresponds to an edge in the mesh, is assigned as a random vector sampled from a Gaussian distribution, and the MeshCNN is trained to deform the mesh by minimizing the Chamfer distance to the input point cloud.
Once converged, the mesh topology is updated by reconstructing a new watertight mesh with more vertices using methods in \cite{huang2018robust}, which is then refined again using MeshCNN.

{We apply Chamfer distance loss (Eq. \ref{eq:chamfer}) and the edge length regularization (Eq. \ref{eq:edge_length}) for the 3D-Prior network. The loss function is defined as $L=\lambda_0 L_{chamfer} + \lambda_1 L_{edge}$, where $\lambda_0$ and $\lambda_1$ balance the loss term and are empirically set to 1.0 and 0.2 respectively in our experiment. $L_{chamfer}$ and $L_{edge}$ are calculated with following equations:
\begin{equation}
    L_{chamfer}=\sum_{\hat{p}} \min_{q}\|\hat{p}-q\|_{2}^{2}+\sum_{q} \min_{\hat{p}}\|\hat{p}-q\|_{2}^{2},
    \label{eq:chamfer}
\end{equation}
\begin{equation}
    L_{edge}=\sum_{p} \sum_{k \in \mathcal{N}(p)}\|p-k\|_{2}^{2},
    \label{eq:edge_length}
\end{equation}
where $p$ is a vertex of the generated mesh; $\hat{p}$ is a point sampled from the generated mesh surface; $q$ is a point in the input point cloud;  $\mathcal{N}(p)$ is the one-ring neighboring vertex of $p$ in the generated mesh. Note that the loss function is slightly different from that in Point2Mesh. We find the beam-gap loss in Point2Mesh is computational expensive with limited effect for our system. Meanwhile, adding the regularization term for edge length speeds up the convergence of our network in practice.}

\subsection{2D Prior Network}
We then refine the mesh via a 2D prior network,
The core idea is to create a 2D representation of the 3D mesh, which can be refined effectively with strong self-prior in 2D CNN.

\subsubsection{Creating XYZ Map}
In this section, we introduce how to create a 2D representation of the 3D mesh which can be refined via 2D CNN.
We first create a UV atlas from the initial mesh using OptCuts~\cite{optcut2018}, while theoretically any atlas generation method respecting UV space continuity would also work \xk{and more discussions are provided in the supplemental material.}
Then, for each point $q$ in the input point cloud, we find its nearest $\hat{p}$ on the initial mesh.
We then query the triangle face ID and barycentric coordinate within the triangle for $\hat{p}$ to calculate a $uv$ coordinate in the texture atlas, at which the $(x,y,z)$ of $q$ is assigned.
Eventually, a sparse three-dimensional XYZ map in the UV space is created to record the 3D locations of all the points in the input point cloud as shown in Fig. \ref{fig:overview}.
We also try to build a distance UV map w.r.t. an anchor point but find it less effective than the XYZ map.

\subsubsection{XYZ Map Refinement}
\label{sec:2dprior}
We then train a network to produce a dense XYZ map supervised by the sparse one.
We adopt a 2D U-Net with skip-connections following DIP~\cite{ulyanov2018deep}.
The network takes as input a random noise feature map $z+\epsilon$, and the weights of the network are randomly initialized.
$z$ is a $32\times Height \times Width$ random vector sampled from Gaussian distribution $\mathcal{N}(0,0.1)$, which is fixed during training. $\epsilon$ is a small Gaussian permutation $\mathcal{N}(0,0.02)$ added to $z$ to prevent the network from overfitting, which changes at every forwarding pass. 
The network is supervised by the sparse XYZ map using $L_2$ loss, and produces the dense XYZ map, whose architecture can be found in the supplementary material.

Note that the 2D U-Net predicts $(x,y,z)$ at each integer pixel location on UV atlas, while $uv$ coordinates of the input points are usually floating points.
To obtain accurate supervision from the sparse ``ground truth'' XYZ map, we use the differentiable bilinear sampling to obtain the value from the predicted dense XYZ map on the sub-pixel locations that have ground truth.

\begin{figure}[!t]
\centering \includegraphics[width=1.0\linewidth]{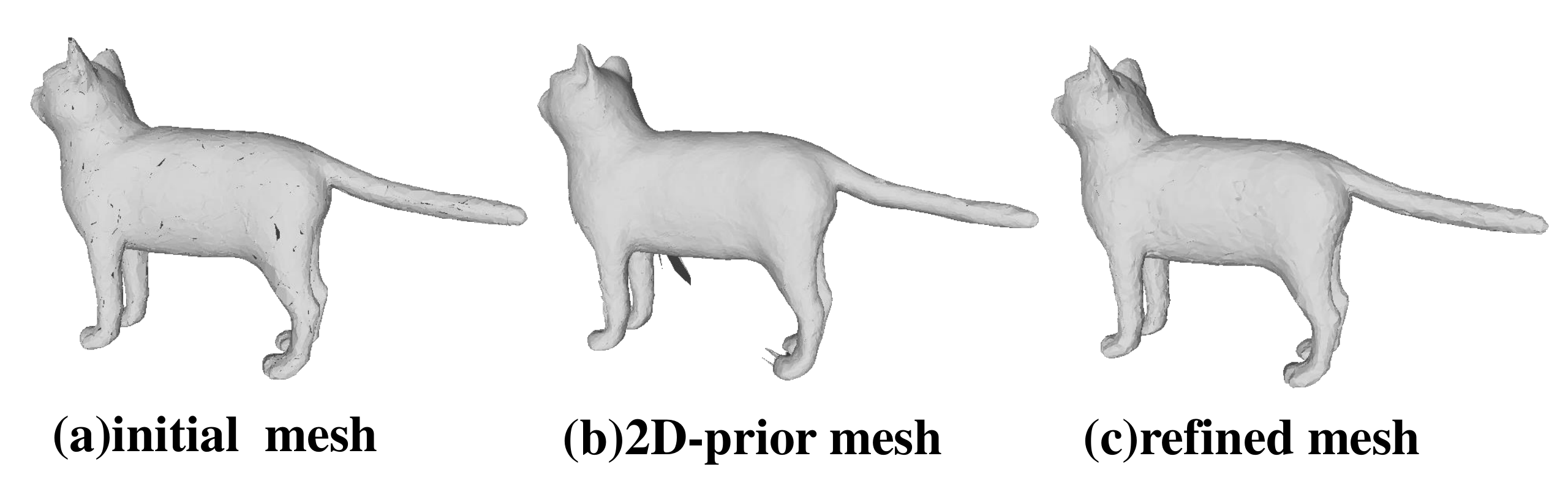}
\vspace{-8mm}
\caption{Iterative Refinement with 2D-3D prior networks.}
\vspace{-4mm}
\label{fig:why-3d} 
\end{figure}

With the dense XYZ map from the 2D-prior network, we update the 3D mesh directly by updating the vertex locations to the value in the predicted dense XYZ map.
Fig. \ref{fig:why-3d} shows an example. Compared to the initial mesh (Fig.\ref{fig:why-3d} (a)), {the 2D-prior refined mesh (Fig.\ref{fig:why-3d} (b)) is smoother with a few spikes, and folded regions are fixed.} 

\subsection{Iterative Refinement with 2D and 3D Priors}
The refinement via the XYZ map significantly improves the majority part of the 3D geometry \xk{and enforces the smoothness}.
However, some obvious artifacts show up, which typically happens on the locations mapped to boundary of valid regions in UV atlas. A possible reason is that the 2D-prior network is weak at completing region without any supervision signal, e.g. the invalid region on UV atlas, and errors are propagated to the few pixels on the boundary of the valid region,  \xk{introducing flipped faces and outlier vertexes}. 
We try to expand the UV atlas valid region for a few pixels as commonly adopted by many previous methods but find this not effective since the supervision is too sparse.
To fix the problem, we send the updated mesh back to the 3D-prior network for another refinement,
and find the 3D-prior network is very effective in keeping benefits from the 2D-prior network and removing the artifacts (Fig.\ref{fig:why-3d} (c)) \xk{via
stronger supervision and regularization}.
The two networks can run iteratively to improve surface quality gradually.

\subsection{Texture Reconstruction}
While most of the previous works only on surface reconstruction, our model can also produce a high-resolution texture, which particularly complements commodity 3D scanners since texture map and color images are usually not provided.
Thanks to the 2D-prior network, we encode the $(r,g,b)$ of the input point cloud into a sparse UV map instead of the location $(x,y,z)$, and run the 2D-prior network to reconstruct a dense texture map supervised by the sparse signal from the input point cloud.
Note that it is also possible to directly produce color for each point in the MeshCNN framework, but the resolution, i.e. color from roughly 10K points, is far from enough to deliver visual appealing rendering quality, e.g., 480K pixels for even VGA resolution.

\subsection{Implementation Details}
The GPU memory required for MeshCNN optimizations is linearly increased as the mesh resolution increases. Following Point2Mesh~\cite{hanocka2020point2mesh}, we cut the mesh into parts to guarantee that each part has less than 6000 faces. There are overlapping regions between different parts, and the final vertex position output is the average over all overlaps.

The whole architecture is implemented in Pytorch, and optimized using Adam Optimizer~\cite{kingma2014adam}. 
For 3D-prior network, the initial mesh contains 2000 vertices.
In each refinement iteration, the 3D-prior network is optimized for 2000 steps with a learning rate of 1e-3, and the 2D-prior network is optimized for 4000 steps.
The mesh is refined for 3 iterations as the performance usually saturates.
The overall generation process takes roughly 2.5 hours, where \xk{3D prior initialization, 2D prior refinement,
3D prior refinement take 90, 20, 40 minutes respectively}. {The 2D-prior network takes much less time compared to 3D-prior network. The runtime of the 2D-prior network is relatively constant w.r.t the number faces, while run-time of the 3D-prior network grows roughly linearly w.r.t the face number. Meanwhile, our initialization stage is faster than Point2Mesh which takes more than 3 hours.}

\begin{figure*}[htb]
\centering \includegraphics[width=1\linewidth,trim={0 1.3cm 0 0}, clip=true]{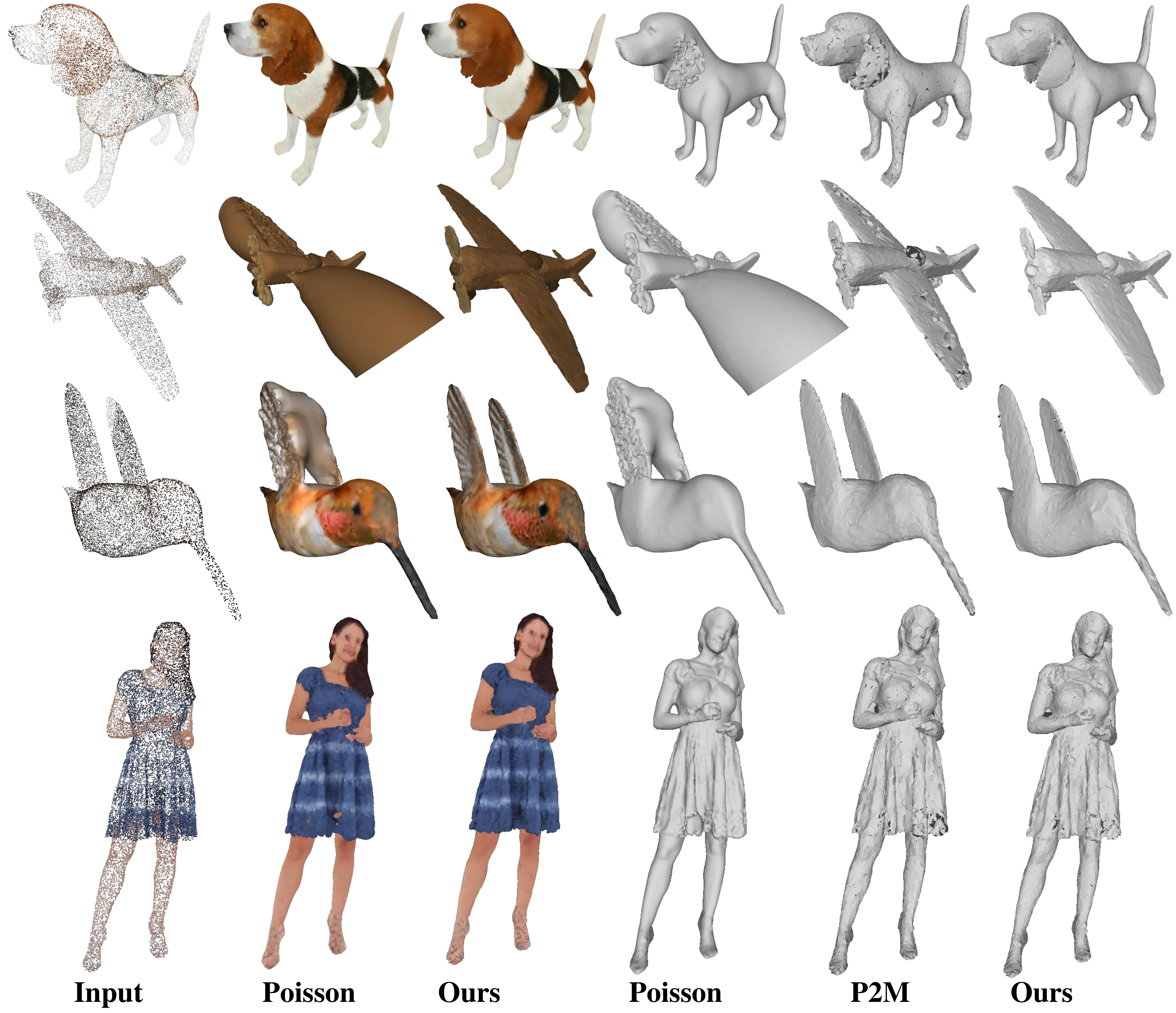}
\vspace{-4mm}
    \begin{tabularx}{0.95\linewidth}{c *6{>{\Centering}X}}
    \hspace*{1cm}Input &\hspace*{0.8cm} Poisson &\hspace*{1cm} Ours & \hspace*{0.4cm} Poisson & \hspace*{0.4cm} Point2Mesh & Ours
    \end{tabularx}
\caption{Comparison on synthetic data. \xk{We show examples with a full spectrum of difficulties for both geometry and texture, e.g. dog and airplane have more complex geometry, and bird has challenging texture. Please refer to our supplementary material for the comparison with more methods. 
}}
\vspace{-4mm}
\label{fig:results} 
\end{figure*}

\begin{figure*}[htb]
\centering \includegraphics[width=1\linewidth,trim={0 2.3cm 0 0}, clip=true]{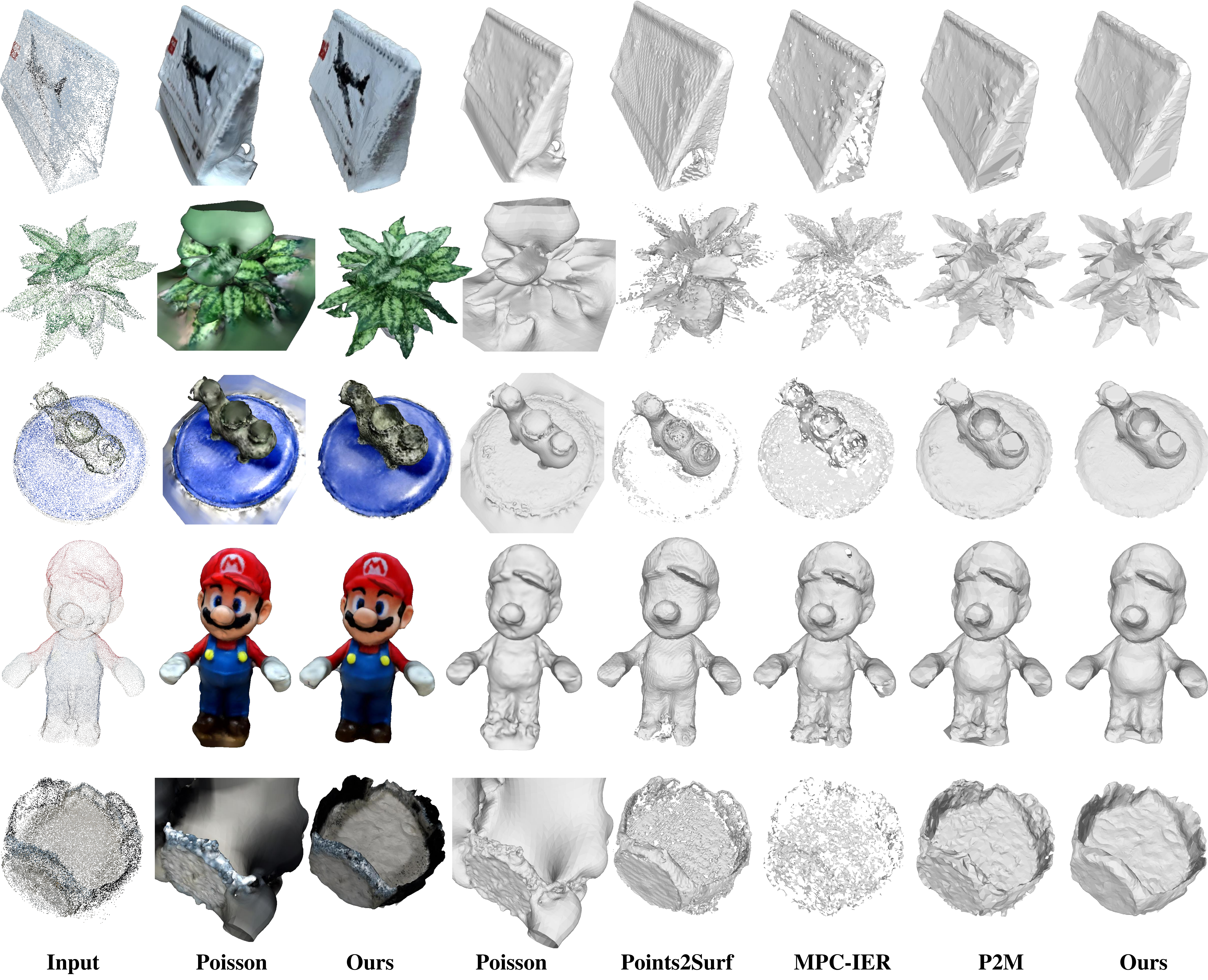}
\vspace{-4mm}

    \begin{tabularx}{1\linewidth}{c *8{>{\Centering}X}}
    \hspace*{0.5cm}Input &\hspace*{0.3cm} Poisson &\hspace*{0.5cm} Ours &\hspace*{0.5cm}  Poisson & \hspace*{0.2cm} Points2Surf & MPC-IER & Point2Mesh & Ours
    \end{tabularx}
    \vspace{-7mm}
\caption{Comparison between our method and other surface reconstruction methods with real scans.}

\label{fig:scan} 
\vspace{-4mm}
\end{figure*}
\begin{figure}[htb]
\centering \includegraphics[width=0.99\linewidth,trim={0 2.5cm 0 0}, clip=true]{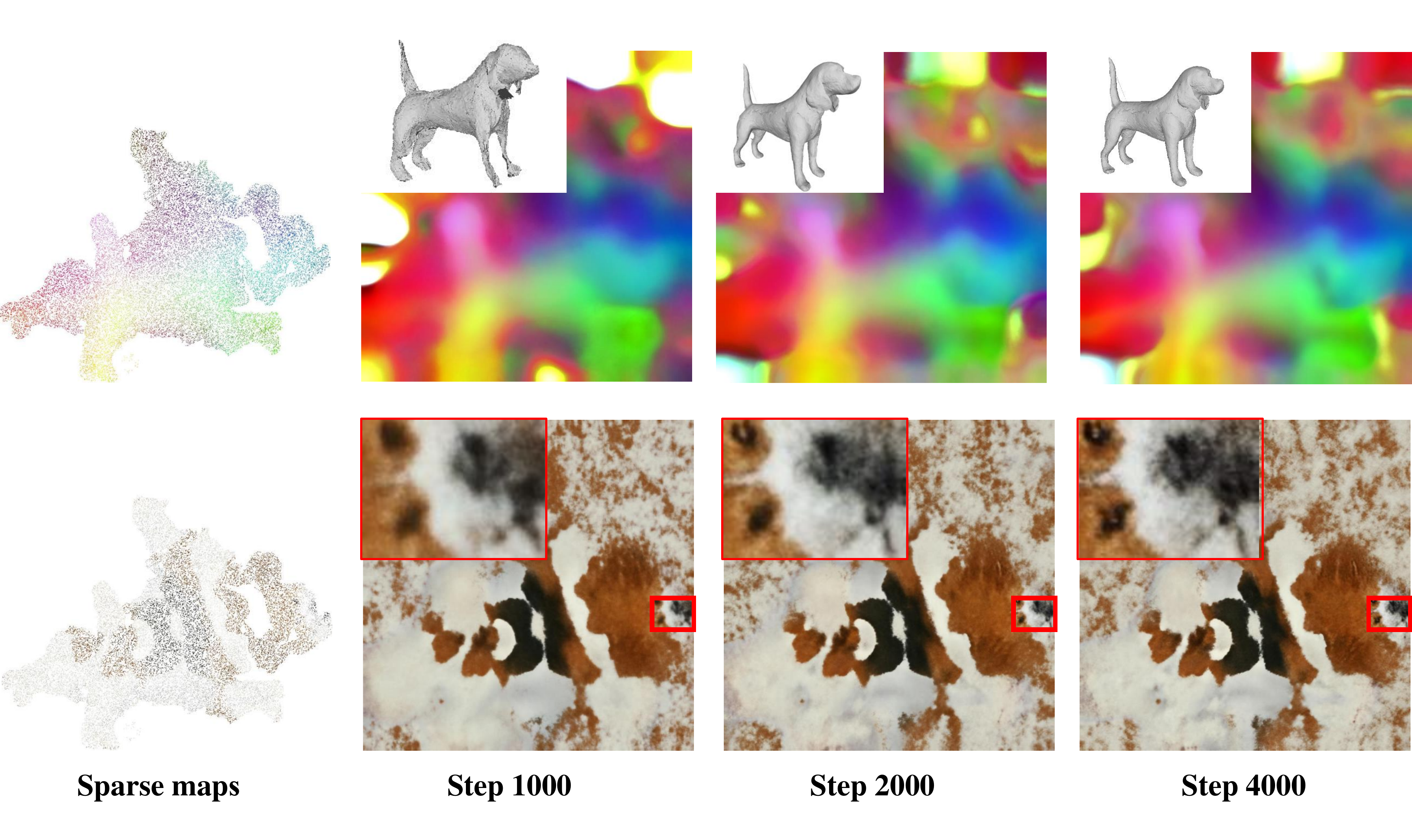}
\vspace{-4mm}
    \begin{tabularx}{1\linewidth}{c *4{>{\Centering}X}}
Sparse maps& Step 1000& Step 2000 &Step 4000
    \end{tabularx}
\caption{Sparse maps and generated dense maps at different training steps.}
\vspace{-4mm}
\label{fig:step} 
\end{figure}

\section{Experiments}
In this section, we evaluate our method for full 3D mesh model generation. We first show overall performance of our system, and then evaluate the quality of geometry and texture separately.
We highlight the effectiveness of our 2D network for both geometry and texture reconstruction.

\subsection{Data and Evaluation Metrics}

{We collected 14 ground-truth meshes for evaluation, including animals, persons, and planes from 3D model repository turbosquid\footnote{\url{https://www.turbosquid.com}} and free3D\footnote{\url{https://free3d.com}}.} All models are 3D meshes with high-quality textures, and the amount of vertices in the original mesh varies from 8000 to 50000. For fair comparison with Point2Mesh~\cite{hanocka2020point2mesh}, we did our best to \xk{collect a similar testing set as ~\cite{hanocka2020point2mesh}, which contains 4 meshes from Thingi10K~\cite{zhou2016thingi10k}, 18 meshes from COSEG~\cite{wang2012active}, and 10 meshes from ShapeNet~\cite{ChangFGHHLSSSSX15} of various object categories.\footnote{\xk{Note that the exact testing set of Point2Mesh is not publicly released.}} }
We synthetically generate input point clouds by uniformly sampling the mesh surface with color.
Each point cloud has 25000 colored points, except for the rabbit which has 10000 points because its structure is relatively simple.

To evaluate the accuracy of the reconstructed model,
we use the F-score as \cite{hanocka2020point2mesh}. {We also show quantitative comparison on Chamfer Distance, Normal Consistency and Earth Mover's Distance(EMD).}
For calculation, we sample 500K points on the surface of the ground truth and the predicted mesh respectively, and empirically find these are sufficiently dense to calculate stable metrics.
For F-score, we set the distance threshold at 0.1 with the span of the longest dimension of each model scaled to 100.

\begin{table}[tb]
    \begin{centering}
    \begin{tabular}{ccccc}
\toprule 
 & F-score$\uparrow$ & CD$\downarrow$ & EMD$\downarrow$ & NC$\uparrow$\\
\midrule 
Poisson  & 95.6 & 0.0630 & 0.1285 & 0.908\\
P2M  & 97.2 &  0.0601 & 0.1068 & 0.941\\
MPC-IER  & 93.2 &  0.0712 & 0.1044 & 0.933\\
Points2Surf  & 90.4 &  0.0974 & 0.1191 & 0.903\\
Ours & \textbf{97.7} &  \textbf{0.0526} & \textbf{0.0969} & \textbf{0.956} \\
\bottomrule 
\end{tabular}
\vspace{-2mm}
    \caption{\label{tab:geometry}Comparison between our method and other surface reconstruction methods on synthetic data. \textbf{Bold}: Best. CD: Chamfer Distance. NC: Normal Consistency.}
    \vspace{-4mm}
\end{centering}
\end{table}

\begin{table}[h]

    \centering
\renewcommand\tabcolsep{4pt}
\begin{tabular}{cccccc}
\toprule 
Dataset & Method & F-score$\uparrow$ & CD$\downarrow$ & EMD$\downarrow$ & NC$\uparrow$\tabularnewline
\hline 
\multirow{2}{*}{Thingi10k} & P2M & 93.4 & 0.0428 & 0.1220 & 0.841\tabularnewline
 & Ours & \textbf{96.1} & \textbf{0.0375 } & \textbf{0.1126} & \textbf{0.884}\tabularnewline
\hline 

\multirow{2}{*}{COSEG} & P2M & 92.8 & 0.0422 & 0.1327 & 0.925\tabularnewline
 & Ours & \textbf{96.2} & \textbf{0.0349} & \textbf{0.1083} & \textbf{0.954} \tabularnewline
\hline 
\multirow{2}{*}{ShapeNet} & P2M & 91.4 & 0.0533 & 0.1417 & 0.895 \tabularnewline
 & Ours & \textbf{95.7} & \textbf{0.0405} & \textbf{0.1164} & \textbf{0.936}  \tabularnewline
\bottomrule

\end{tabular}
   \caption{Comparison with Point2Mesh. \textbf{Bold}: Best. CD: Chamfer Distance. NC: Normal Consistency.}
   \label{tab:thingi10k}
    \vspace{-4mm}
\end{table}
\subsection{Full 3D Mesh Generation}
We first show the quality of textured mesh generated by our model.
Fig.~\ref{fig:results} shows the input colored point cloud and the generated mesh visualized with and without the generated texture.
Overall, our method successfully recovers thin geometry, e.g. the dog's ear and the bird's wings, and texture with sharp boundaries and details.
As a comparison, we show the full model generated by Poisson reconstruction \cite{kazhdan2013screened} {where the surface normal is calculated using implementations in MeshLab~\cite{cignoni2008meshlab} for well-known accuracy and robustness}, and the texture is generated by linearly blending the color from the input point cloud \cite{kazhdan2019adaptive}. 
From Fig.~\ref{fig:results}, our method clearly outperforms others on the quality of both geometry and texture. 

We visualize the sparse XYZ map and the texture map and the dense predictions from our method in Fig. \ref{fig:step}.
We increase the size of the point (4 pixels for each point) for visualization since the original input is too sparse to see.
Even with such sparse input, the network still manage to produce dense output and improve the details over iterations.

\subsection{Surface Reconstruction}
\label{sec:surface}
In this section, we evaluate the quantitative accuracy of the surface reconstruction, and the results are shown in Tab. \ref{tab:geometry}. In addition to Poisson surface reconstruction~\cite{kazhdan2013screened} and Point2Mesh~\cite{hanocka2020point2mesh}, we also compare to Points2Surf~\cite{erler2020points2surf} and MPC-IER~\cite{liu2020meshing} for surface reconstruction.
Note that they both require additional 3D data for training while we do not but purely rely on self-prior.
Our method achieves the best score among all the methods, indicating our geometry is more accurate compared to other methods. {Our method also shows consistently better performance than Point2Mesh on Thingi10K, COSEG and ShapeNet as shown in Tab.~\ref{tab:thingi10k}.} Note that for surface reconstruction, Point2Mesh can be considered as an ablation of our method without the 2D self-prior.
Therefore, our improvements are mostly benefited from the use of the proposed hybrid 2D-3D prior. \xk{We also show the performance w.r.t iteration in the supplementary material.}

The qualitative results are shown in Fig. \ref{fig:results}.
{We can see that Poisson reconstruction generates incorrect meshes for thin structures, e.g., wings of the bird and plane. The results of Point2Mesh are relatively noisy, e.g., the dog's ear and the plane's wings, due to its weak 3D self-prior. In contrast, our method has the best performance with smooth meshes and correct thin structure.}

\begin{figure}[htb]
\centering \includegraphics[width=1\linewidth,trim={0 2.5cm 0 0}, clip=true]{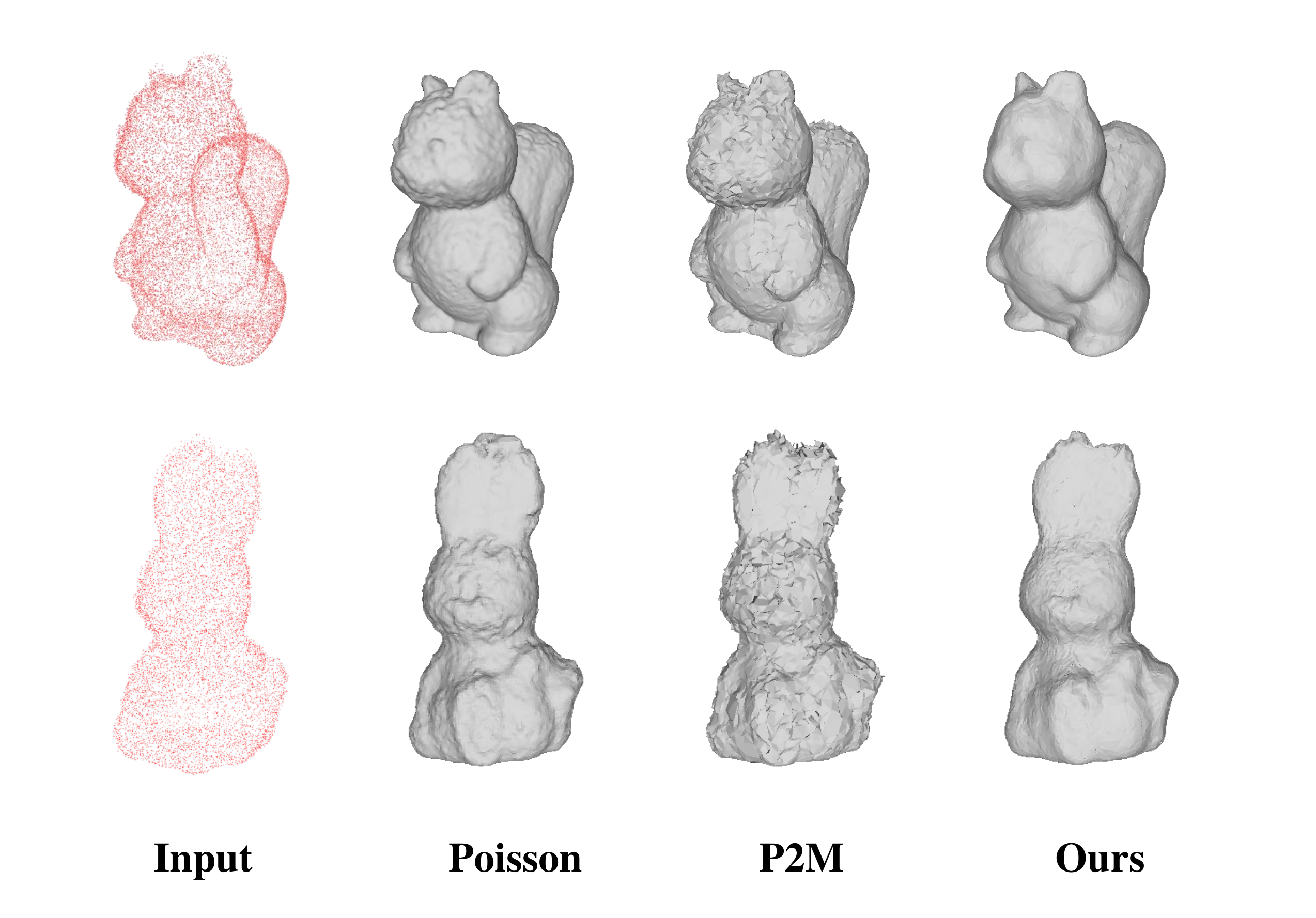}
\vspace{-4mm}
\begin{tabularx}{0.8\linewidth}{c *4{>{\Centering}X}}
    \hspace*{0.1cm}Input &     \hspace*{0.2cm}Poisson & \hspace*{0.2cm} P2M & \hspace*{0.2cm} Ours
\end{tabularx}
\vspace{0mm}
\caption{Comparison between our method and other surface reconstruction methods with noisy input.}
\vspace{-4mm}
\label{fig:noise} 
\end{figure}

\paragraph{Robustness Against Noise}
It is well-known that the quality of the surface reconstruction highly depends on the input point cloud quality.
We test the robustness of our method by manually adding Gaussian noise with standard deviation as $2\%$ of the original coordinate value on the input point cloud, and the results are shown in Fig. \ref{fig:noise}.
Both Poisson and Point2Mesh are easily get affected by the noise in the input, and the surface quality is inferior compared to ours.
Especially, more noise is reflected in the results from Point2Mesh, which indicates the self-prior is not strong enough in 3D GCN.
Comparatively, our method leveraging hybrid 2D-3D self-prior still produce reasonable surface quality by removing noise in the XYZ map. \xk{More results w.r.t different noise levels and robustness on texture generation are
provided in the supplementary material.}
\begin{figure}[tb]\centering      
\subfigure[F-score]{                    
    \label{fig:fscore}           
\includegraphics[scale=0.25]{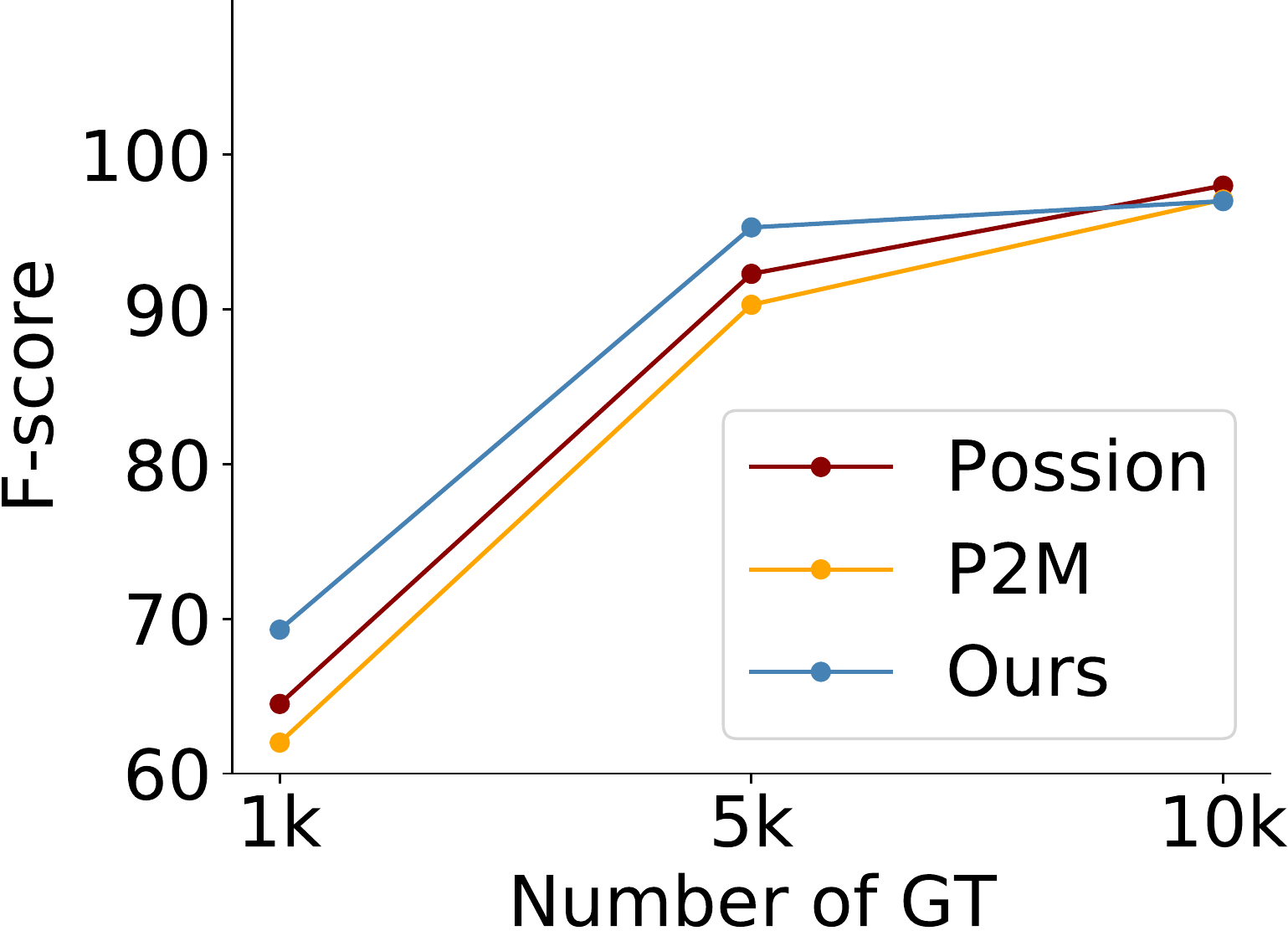}

}                                  \subfigure[Chamfer]{                    
\label{fig:chamfer}            
\includegraphics[scale=0.25]{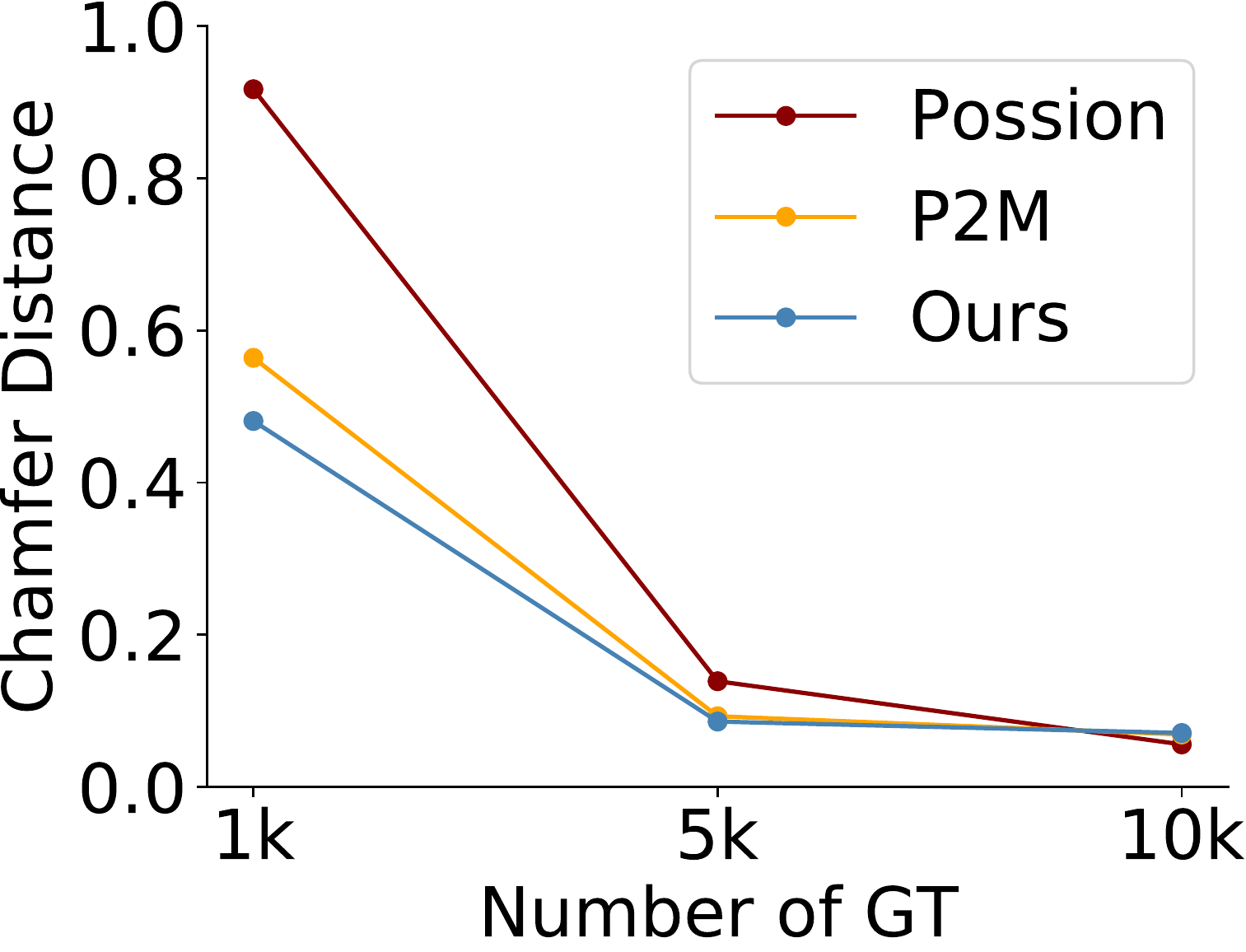}}
\vspace{-4mm}
\caption{Comparison with Screen Poisson reconstruction and Point2Mesh with different sparseness. (a) F-score metric. Higher is better. (b) Chamfer metric. Lower is better.}
\label{fig:sparse_com} 
\vspace{-4mm}
\end{figure}
\paragraph{Robustness Against Sparsity}
We also test the robustness of our model on sparse input point clouds. We run Poisson, Point2Mesh, and our method by taking input point clouds with different number of points, and show the performance in Fig. \ref{fig:sparse_com}.
As expected, the performance of all the methods drops when the input point number is decreasing, but our performance drops slower than the others, which again shows that the 2D-3D self-prior is strong to interpolate relatively large missing regions. More qualitative results are provided in the supplementary material.

\begin{figure}[htb]
\centering \includegraphics[width=0.95\linewidth]{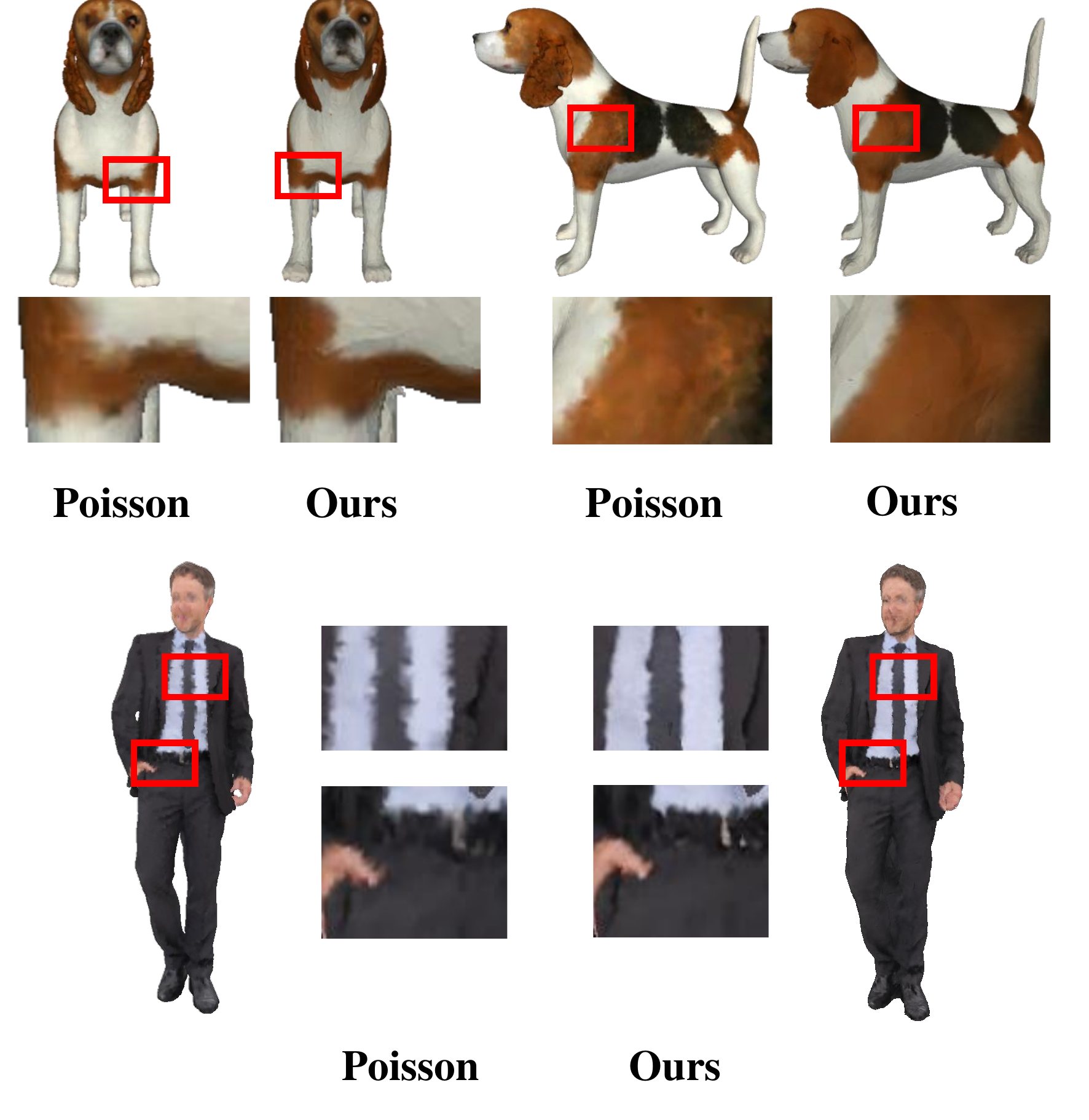}
\vspace{-4mm}
\caption{Comparison between our method and the Poisson surface reconstruction on texture quality. Our method produces texture with better quality especially on edges.}
\vspace{-4mm}
\label{fig:detail} 
\end{figure}

\subsection{Texture Reconstruction}
In this section, we evaluate our texture reconstruction.
It is not easy to find previous work with code for this specific task under our setting, so we compare to Kazhdan  \etal ~\cite{kazhdan2019adaptive} which is implemented with Poisson surface reconstruction in Meshlab. 
In Fig. \ref{fig:detail}, we show the texture on the reconstructed mesh and highlight some regions.
Our method significantly outperforms the baseline method especially on edges, where the baseline texture tends to be blurry and ours is usually sharp. {We also build a MeshCNN baseline method which directly predicts color for each point in the MeshCNN~\cite{hanocka2019meshcnn} framework and provided the comparisons in supplementary material.} 

{For the quantitative measure, we render the generated colored mesh uniformly in 16 different views as~\cite{martin2020gelato} and evaluate the Naturalness Image Quality Evaluator (NIQE) score~\cite{mittal2012making} of the rendered $4096 \times 4096$ images. NIQE is a no-reference metric that solely considers perceptual quality. The NIQE results of MeshCNN baseline, Poisson and our method are 20.39, 20.65 and {19.35} respectively (lower is better), which indicates ours shows better visual quality. }

\subsection{Generalization to Real Scans}
We also test how our method performs on scans collected from the commodity 3D scanner.
We collect 3D textured point clouds of five objects using Huawei 3D Live Maker, each of which contains roughly 30000 points.
Note that the noise and sparsity of the points are not ideally uniform, and thus these data are more challenging for full mesh reconstruction.
Fig. \ref{fig:scan} shows results on these scans.
Compare to other methods, we produce overall better geometry for smooth surface, sharp boundary, thin structure, and sparse areas.
Our texture is also sharper than Kazhdan \etal~\cite{kazhdan2019adaptive}, e.g., the airplane on calendar, plant, and mario face.

\section{Conclusion}
We propose a method to reconstruct textured mesh from a colored point cloud by leveraging self-prior in deep neural networks. A novel hybrid 2D-3D self-prior is exploited in an iterative way. Based on an initial mesh generation using a 3D convolutional neural network with 3D self-prior, the 2D UV atlas is generated and used to encode both 3D information and color information that can be further refined by 2D CNNs with the self-prior.  Experiments demonstrate the advantages of the proposed method over SOTA methods.

\noindent \textbf{Acknowledgement} This work was supported in part by NSFC Project under Grant 62076067.

{\small
\bibliographystyle{ieee_fullname}
\bibliography{3738}
}

\renewcommand\thesection{\Alph{section}}
\renewcommand\thetable{\Alph{table}}
\renewcommand\thefigure{\Alph{figure}}

\clearpage
\noindent \textbf{\Large Supplementary Material}
\setcounter{section}{0}

\begin{figure*}[htb]
\centering \includegraphics[width=0.98\linewidth]{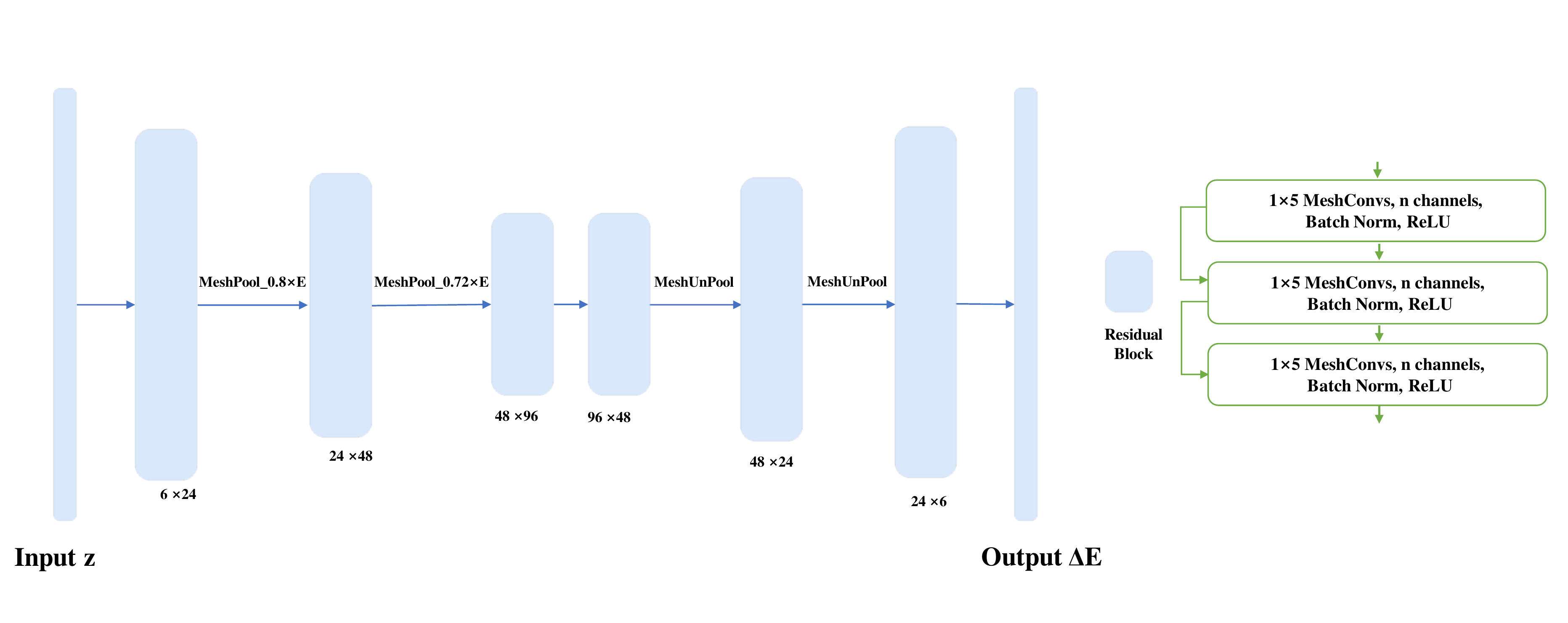}
\caption{Architecture of 3D-prior network. {In general, it consists of six residual blocks. The detailed structure and channels of the residual blocks are shown in the left.}}
\label{fig:arch3d} 
\end{figure*}

\begin{figure*}[htb]
\centering \includegraphics[width=0.98\linewidth]{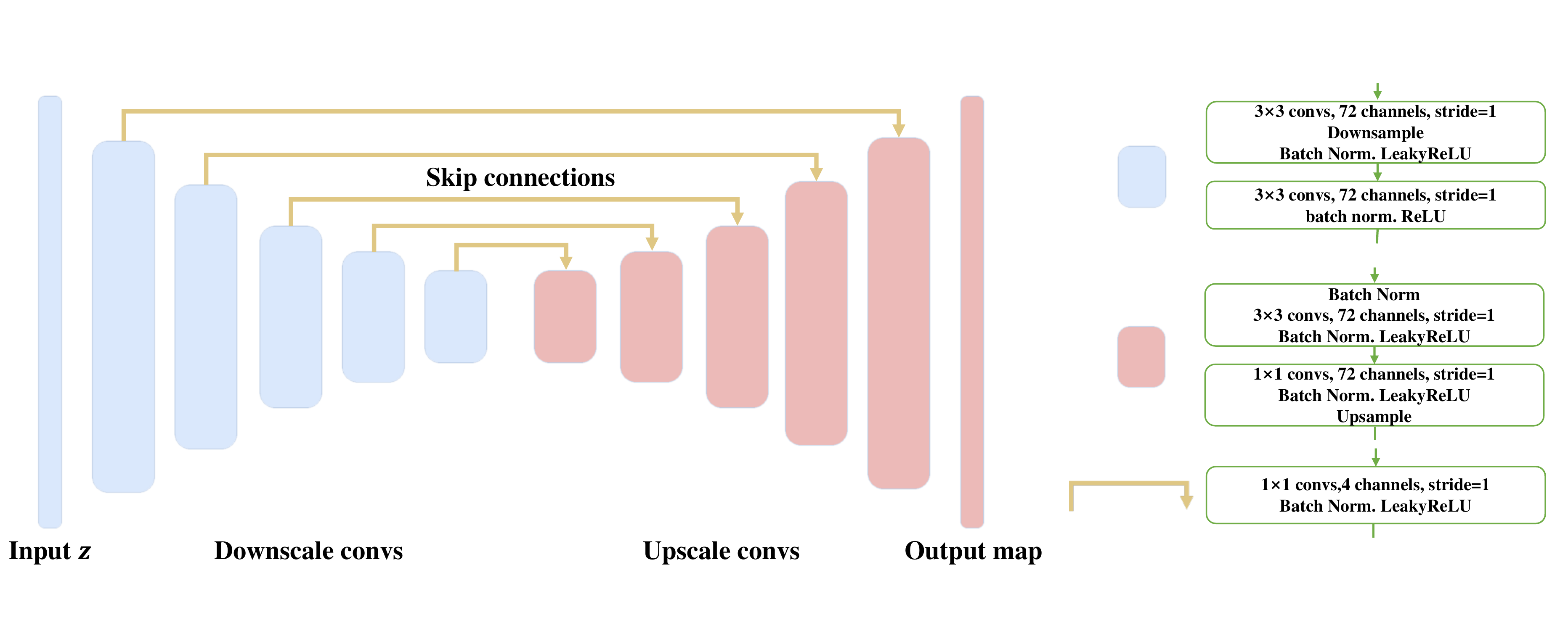}
\caption{Architecture of 2D-prior network. The layers of each component are shown in the right.}
\label{fig:arch2d} 
\end{figure*}

\section{Implementation Details}
In this section, we introduce detailed network architecture for our 2D- and 3D-prior network. Note that both network is initialized with random weights and trained to reconstruct {dense structures from randomly initialized input features with the sparse supervisions}. \xk{Different values of hyperparameters (optimization steps, std. dev. of initialization and loss weights) have been tried, and the method
works for most settings. In practice we choose the hyperpaprameters
for the best trade-off between accuracy and efficiency.}
\subsection{3D-prior Network}
We use a MeshCNN \cite{hanocka2019meshcnn} based  3D-Prior  network  introduced  in  Point2Mesh~\cite{hanocka2020point2mesh} to generate  an  initial  mesh and the refined output mesh.
We also adopt residual and skip connections in MeshConv~\cite{hanocka2019meshcnn} layers which compose a residual block. ReLU is used as the active function after each MeshConv layer except for the last layer.
The network receives an $n_{e} \times 2 \times 3$ dimensional initial random vector $z$ as input {where $n_{e}$ is the number of input edges }, and the network outputs an edge feature vector $\Delta{E}$ with the same dimension that represents the displacement of two vertices on each side of the edges. \xk{In each refinement iteration, the 3D-prior network is optimized for 2000 steps with a learning rate of 1e-3, and the weight of the edge loss is 0.2.}

In Figure~\ref{fig:arch3d}, we show the detailed  architecture  of our 3D-prior network. The whole network includes six residual blocks, two MeshPool layers and two MeshUnpool layers. Each residual block contains three MeshConvs. Input and output channel number of each  residual block and the pool proportion of each MeshPool layer are shown in Figure~\ref{fig:arch3d}.

\subsection{2D-prior Network}
We follow DIP~\cite{ulyanov2018deep} to design our 2D-prior network. An encoder-decoder architecture with several skip-connections is adopted for all of our experiments with same hyper-parameters except for training steps. The number of training steps is 2000 for dense color texture map generation, while for dense XYZ map generation the number is 4000  \xk{with a learning rate of 1e-2.} 
LeakyReLU~\cite{he2015delving} is used as the active function. The downsampling operation in the network is implemented as convolution with strides, and for upsampling we use bilinear upsampling. In each convolution layer, a reflection padding is used instead of zero padding. The input random feature map and the output dense UV map have the same spatial resolution $1024\times1024$. 

In Figure \ref{fig:arch2d} we provide the details of our 2D-prior network architecture. The whole network contains five downscale convolution blocks, five upscale convolution blocks and five skip connection blocks. The layers and parameters of each  block are shown in the right part of Figure \ref{fig:arch2d}.

\xk{\subsection{UV Flattening}
We use OptCuts\cite{optcut2018} to create a UV atlas from the 3D mesh. The UV flattening via OptCuts may not be ideal and would affect the 2D prior network output, but most of artifacts can be fixed by the 3D prior network with strong supervision and regularization, and UV map will be regenerated with improved geometry afterward. Therefore, we find our method doesn’t need perfect UV-maps at the beginning. UV flattening is challenging for complex geometry, but we only need the UV space to preserve some local smoothness regardless of other criteria, such as distortion and seam lengths.}

\begin{figure*}[htb]
\centering \includegraphics[width=0.8\linewidth, trim={0 2.0cm 0 0}, clip=true]{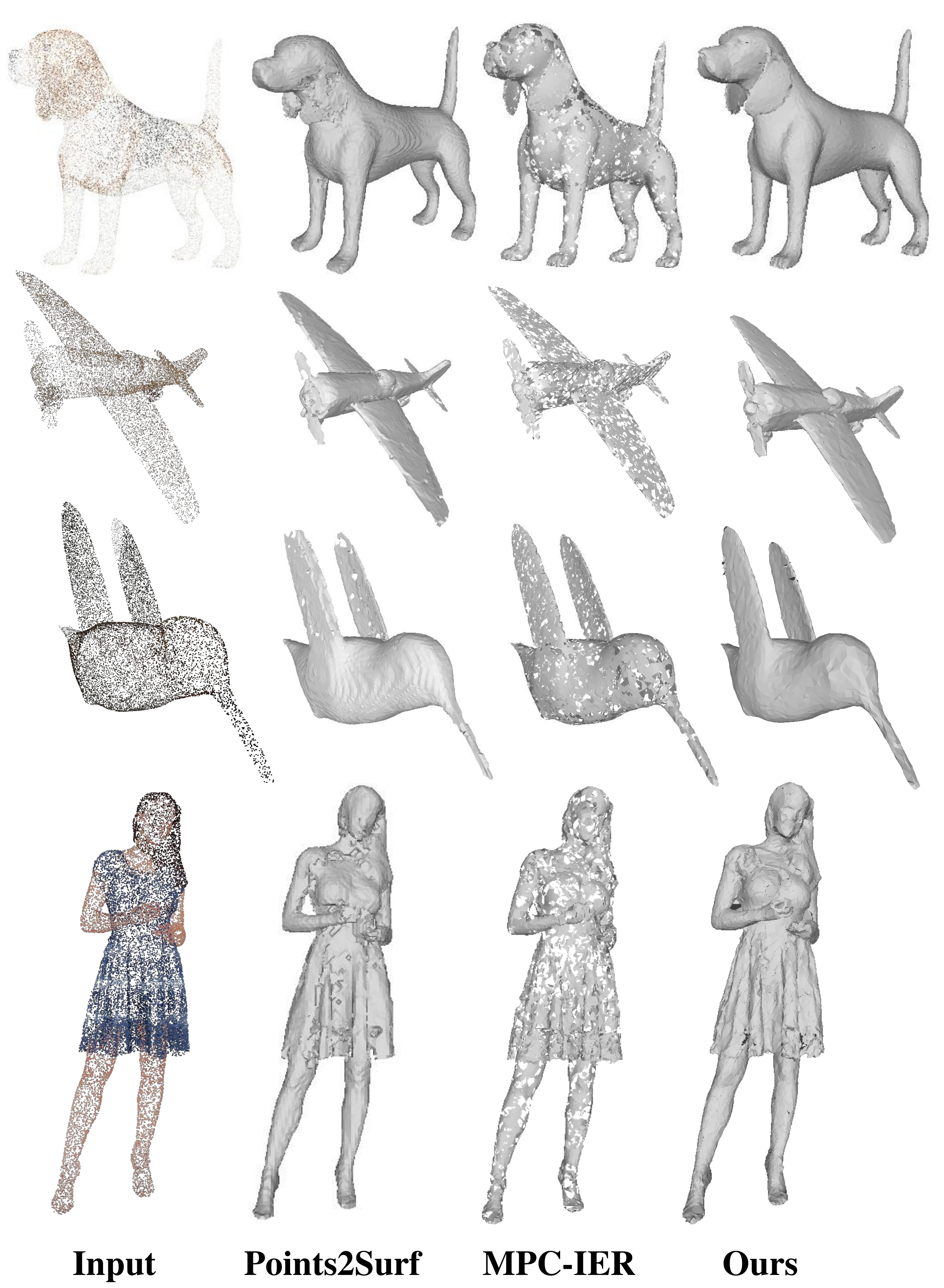}
        \begin{tabularx}{0.95\linewidth}{c *6{>{\Centering}X}} 
    \hspace*{2.5cm}Input &\hspace*{0.5cm} Points2Surf& \hspace*{-1.5cm}  MPC-IER & \hspace*{-4.0cm}Ours 
    \end{tabularx}
\caption{Comparison between our method and more surface reconstruction methods on synthetic data. }
\label{fig:other_res} 
\end{figure*}

\begin{figure*}[htb]
\centering \includegraphics[width=0.99\linewidth]{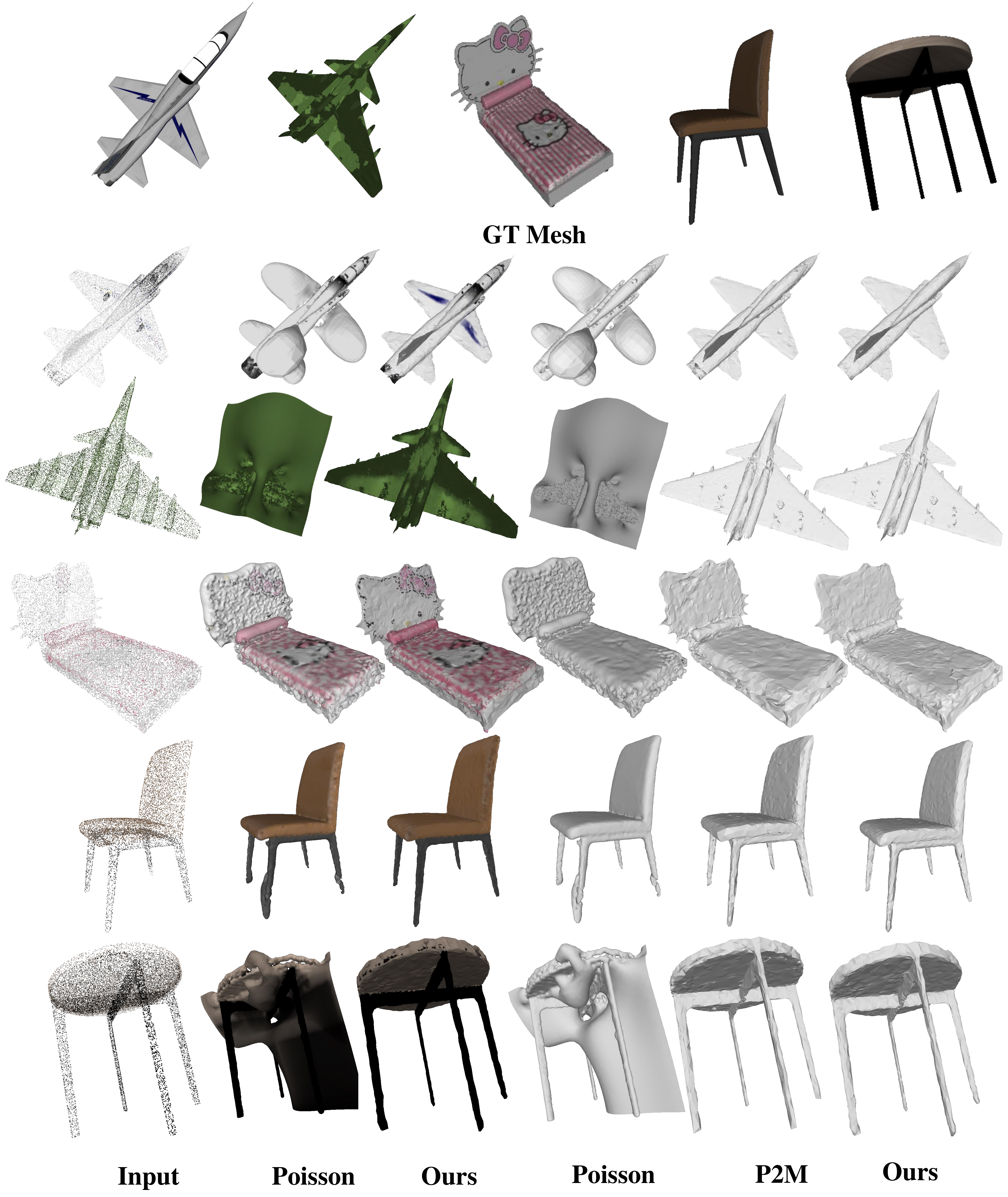}
\caption{Comparison between our method and other surface reconstruction methods. The groundtruth meshes used to sample input point clouds are shown in the first row.}
\label{fig:more_res} 
\end{figure*}
\begin{figure*}[htb]
\centering \includegraphics[width=0.99\linewidth, trim={0 1.65cm 0 0}, clip=true]{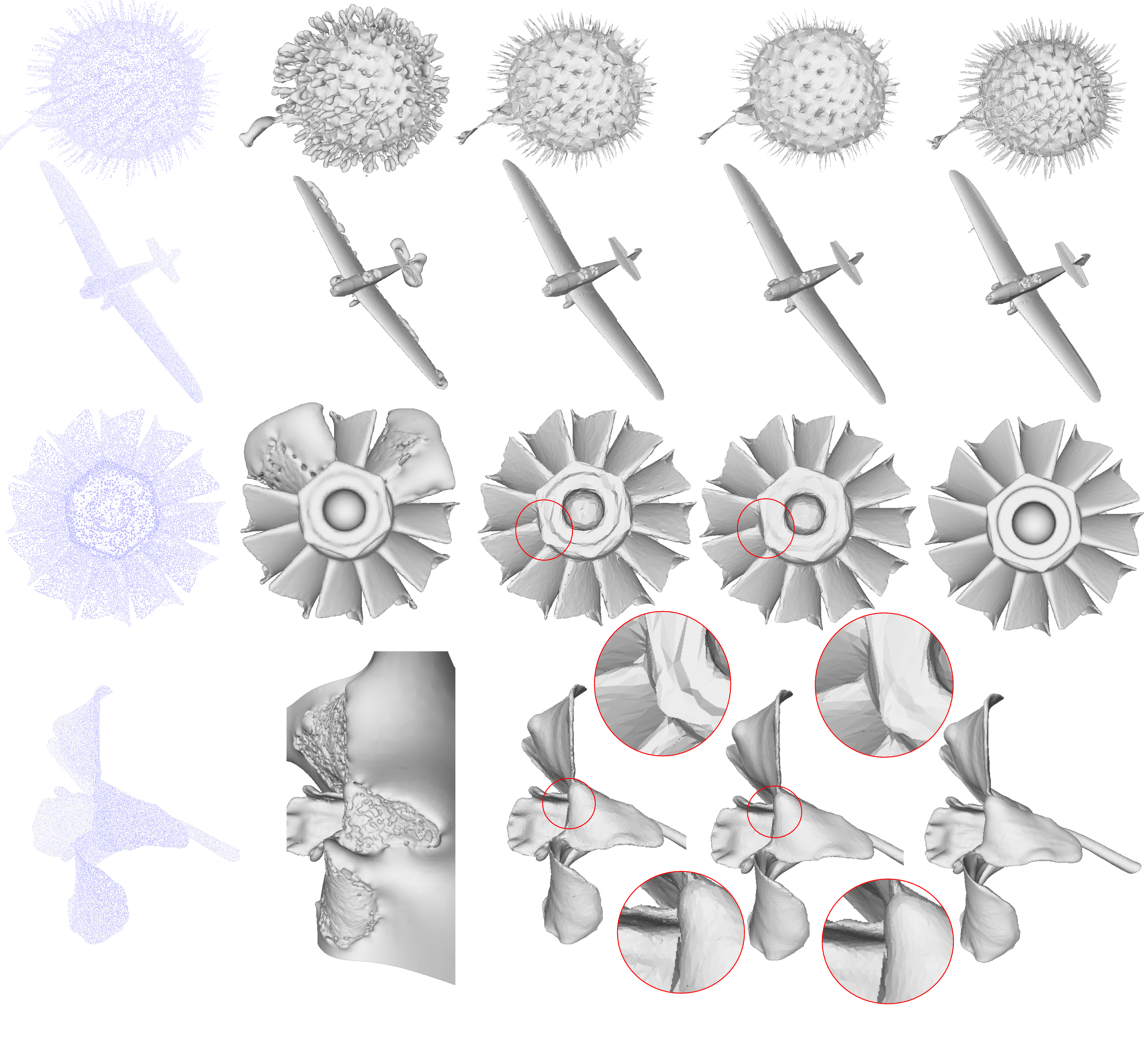}
        \begin{tabularx}{0.95\linewidth}{c *5{>{\Centering}X}} 
    \hspace*{0.5cm}Input &\hspace*{1.9cm} Poisson & \hspace*{1.5cm}  P2M & \hspace*{1.0cm}Ours  & GT
    \end{tabularx}
\vspace{-2mm}
\caption{Comparison between our method and other methods on surface reconstruction.}
\label{fig:more_res_surf} 
\end{figure*}
\section{More Qualitative Results}
In this section, we show more qualitative results in \xk{Fig. \ref{fig:other_res} and} Fig. \ref{fig:more_res}. \xk{In Fig. 4 of main paper, we showed comparison to Poisson~\cite{kazhdan2013screened} and Point2Mesh~\cite{hanocka2020point2mesh} on a few examples. Here we add comparison to more previous works in Fig. \ref{fig:other_res}.} For each example in Fig.~\ref{fig:more_res}, we show the input colored point cloud and the results from Screen Poisson Surface Reconstruction \cite{kazhdan2013screened} (Poisson), Point2Mesh \cite{hanocka2020point2mesh} (P2M),  and our model.
We show the mesh results with and without texture.\xk{ We also show comparison on the surface reconstruction task on the samples with complicate structures in Fig. \ref{fig:more_res_surf}.}

Overall, our model produces shapes and textures that recovers more details and maintains better visual quality. Screen Poisson surface reconstruction tends to make large geometric errors when the points are sparse or {the surface normal cannot be estimated accurately}, such as the wing of the aeroplane and the chair legs. 
The texture map is usually blurry compared to our result generated with hybrid priors.
\begin{figure*}[tb]

\begin{center}

\includegraphics[width=0.99\linewidth, trim={0 0.9cm 0 0}, clip=true]{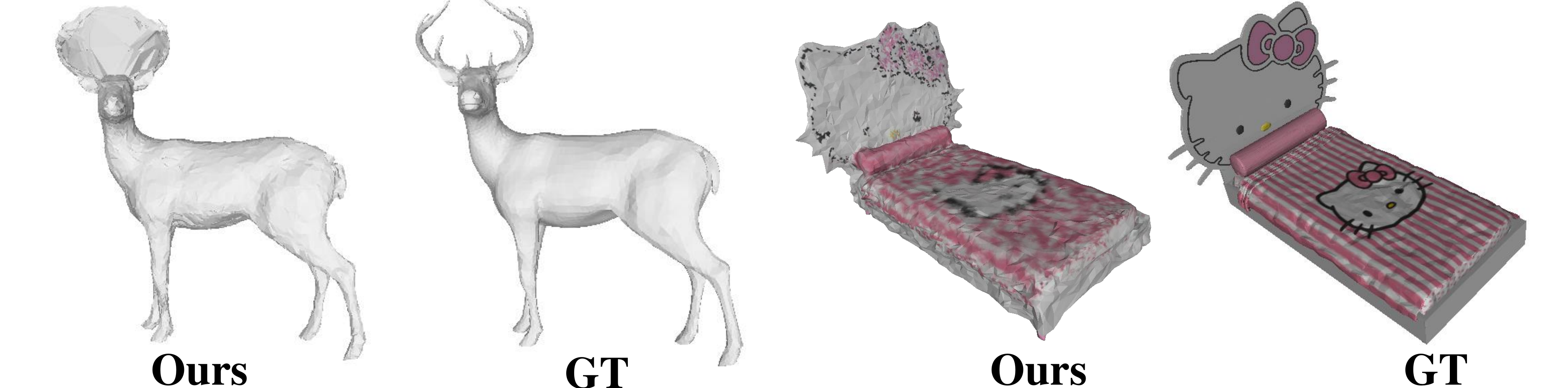}
\end{center}
        \begin{tabularx}{0.95\linewidth}{c *4{>{\Centering}X}} 
     \hspace*{2.0cm}Ours  & \hspace*{2.0cm} GT  & \hspace*{2.0cm}Ours  & \hspace*{1.0cm}GT
    \end{tabularx}
    \vspace{-3mm}
   \caption{\label{fig:rebuttal_f2}Failure cases in geometry and texture reconstruction.}
\vspace{-3mm}
\end{figure*}
\begin{table}[tb]
    \begin{centering}
    \begin{tabular}{ccc}
\toprule 
 & F-score$\uparrow$ & CD$\downarrow$\\
\midrule 
MC-APSS  & 93.4 & 0.0650 \\
MC-RIMLS  & 96.6 &  0.0597 \\
P2M-S & 97.3 & 0.0589 \\
Ours & \textbf{97.7} &  \textbf{0.0526} \\
\bottomrule 
\end{tabular}
    \caption{\label{tab:geometry_more}Comparison between our method and marching cube based methods on synthetic data. The best results are noted by \textbf{Bold}. CD is short for Chamfer Distance.}
    
\end{centering}
\end{table}

\section{Comparison with More Surface Reconstruction Methods}
We show the surface reconstruction comparison with two recent marching cube methods implemented in Meshlab, named MC-APSS\cite{guennebaud2008dynamic} and MC-RIMLS\cite{oztireli2009feature}. \xk{We also show the comparison with Point2Mesh~\cite{hanocka2020point2mesh} with a HC Laplacion smooth~\cite{vollmer1999improved}, namely P2M-S.}
The Chamfer distance and F-Score are reported in Tab.~\ref{tab:geometry_more}, which are worse than our method.

{\section{Failure Cases}
Fig. \ref{fig:rebuttal_f2} shows some typical failure cases of our method on geometry and texture reconstruction. Some local structures are not separated correctly due to taking convex hull as initialization. Meanwhile, our method is unable to recover high frequency strip texture or tiny patterns with unstructured sparse input colored points.
}

\begin{table}[htb]
\centering
\begin{tabular}{ccccc}

\toprule
&  2\% & 5\% & 10\% &\tabularnewline \hline 
Poisson & 77.3 & 53.8 & 25.9 \tabularnewline
P2M &  70.1 & 50.4 & 18.0  \tabularnewline
Ours  & \textbf{83.5}& \textbf{69.3} & \textbf{37.6} \tabularnewline
\bottomrule 

\end{tabular}
     \caption{Comparison on F-score between our method and other surface reconstruction methods with noisy input. The best results are noted by \textbf{Bold}.}
     \label{tab:noise}
\end{table}

\begin{figure}[htb]
\centering \includegraphics[width=0.95\linewidth]{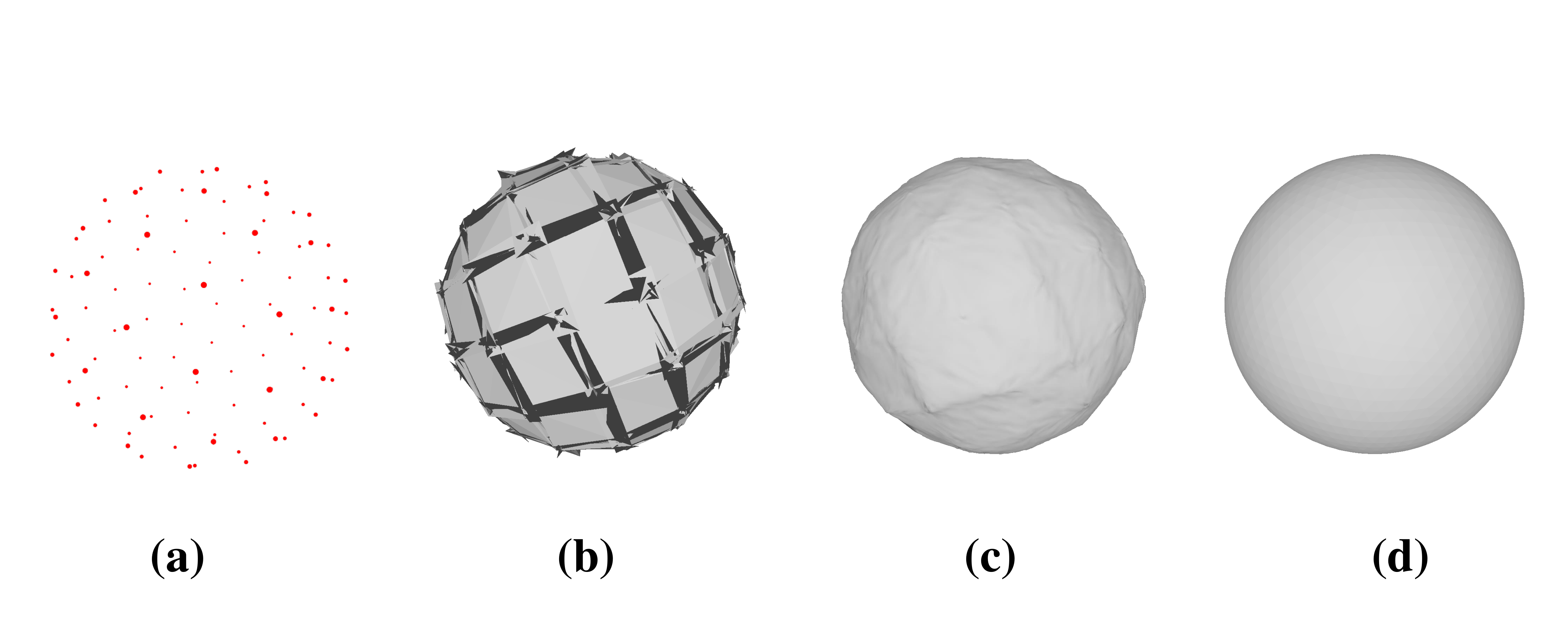}
\caption{Surface reconstruction results of very sparse spherical input. (a) Input point cloud with only 100 points; (b) 3D mesh generated by Point2Mesh; (c) Our result; (d) Ground truth. 
}
\label{fig:sparse_toy} 
\end{figure}

\begin{figure}[htb]
\centering \includegraphics[width=0.95\linewidth]{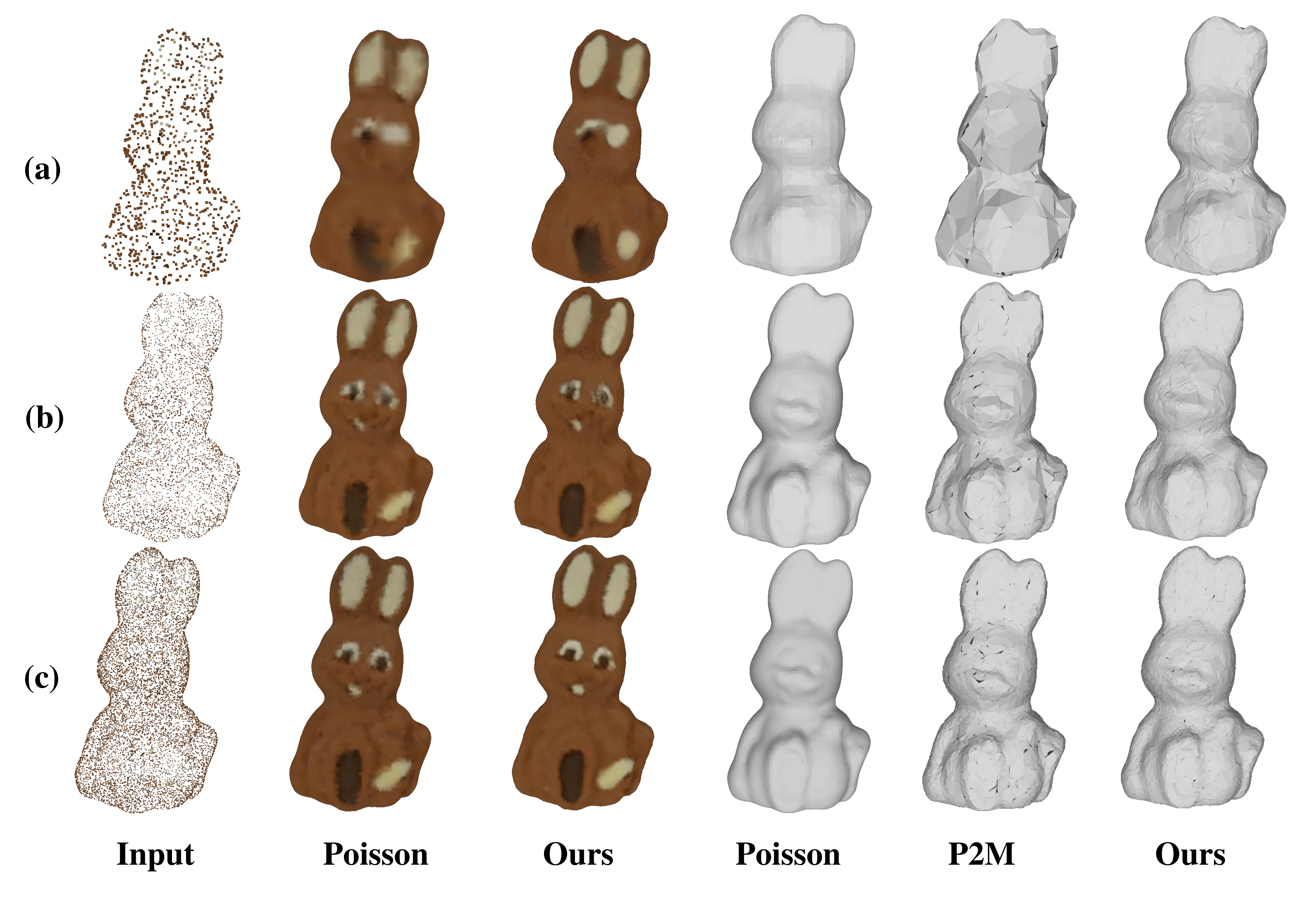}
\caption{Generated textured mesh with different sparseness. (a) 1k input points; (b) 5k input points; (c) 10k input points. }
\label{fig:sparse} 
\end{figure}
\xk{\section{Robustness Evaluation}
\subsection{Robustness against Noise}
We test the robustness of our method by manually adding Gaussian noise to the original coordinate value on the input point cloud. Table~\ref{tab:noise} shows quantitative results as the standard deviation of noise increased from 2\% to 10\%. We also conduct
experiments that add Gaussian noise with standard deviation
as 10\% on color. The NIQE results of MeshCNN
baseline, Poisson and our method are 22.47, 23.90 and
20.16 (lower is better). Compared to the case without noise (see
Sec. 4.4), our method suffers the minimum decline, which
indicates comparatively good robustness.
\subsection{Robustness against Sparsity}}
In Fig. \ref{fig:sparse_toy}, we show a toy example where only 100 points are sampled from the ball as the input for surface reconstruction in order to understand the advantage of our hybrid 2D-3D prior over the 3D only prior in Point2Mesh.
Point2Mesh completely fails since no reasonable prior can be learned by 3D GCN from such an extremely sparse input. In contrast, our method is successful in reconstructing a reasonable ball.
We would like to advocate that this is mostly benefit from the strong prior encoded in the XYZ map. 

Fig. \ref{fig:sparse} shows results with input point clouds of different sparseness on a textured mesh. For Point2Mesh, noisy or large planar surfaces show up quickly when inputs become sparse, while our method still produce smooth surface maintaining roughly correct geometry with certain level of details. 
{With dense input and simple structure, Poisson can generate both good and texture information, while when the input become sparser, it lose more details of texture and geometry compared to our method.} 
\begin{figure*}[htb]
\centering \includegraphics[width=0.96\linewidth, trim={0 1.2cm 0 0}, clip=true]{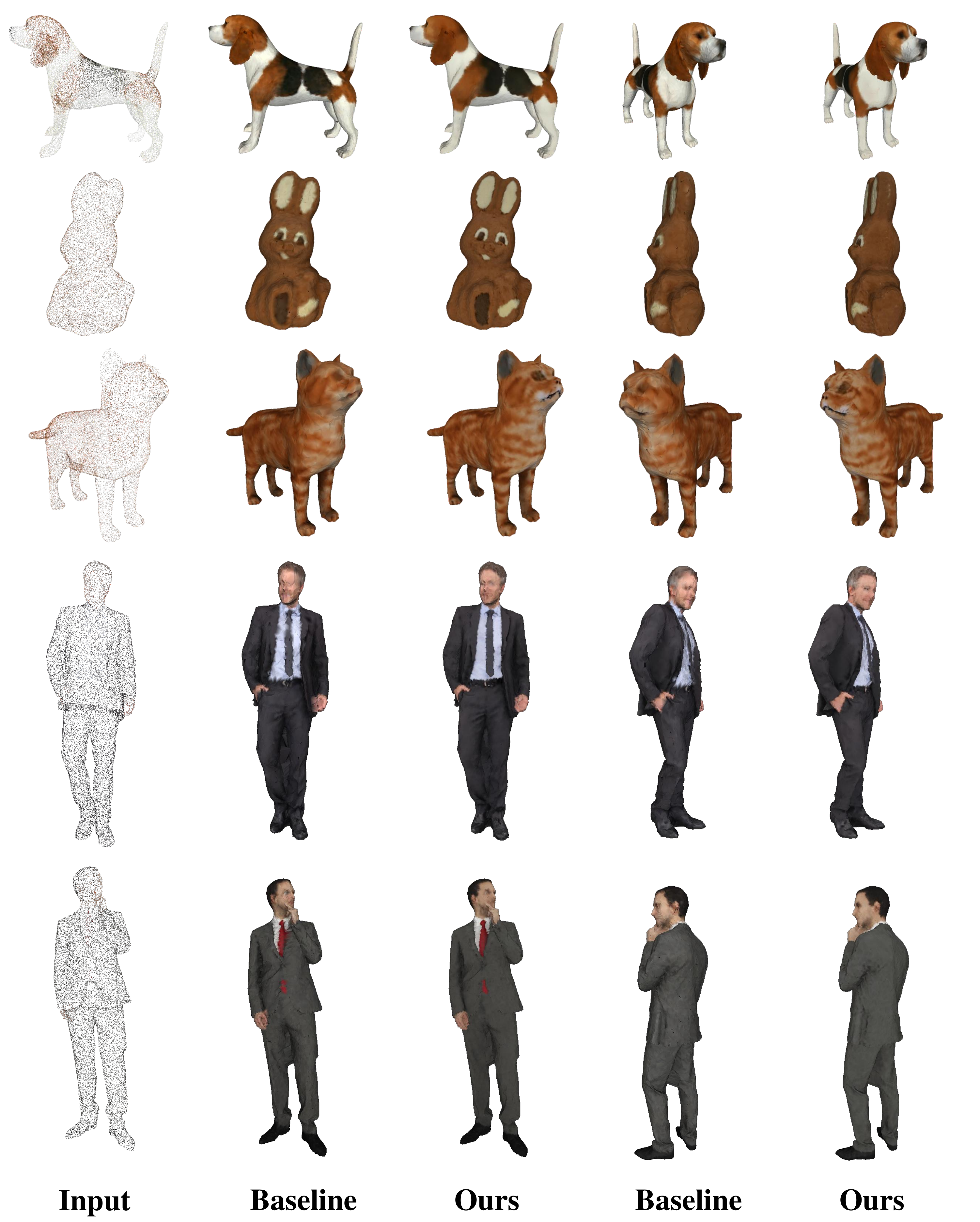}
        \begin{tabularx}{0.95\linewidth}{c *5{>{\Centering}X}} 
    \hspace*{0.8cm}Input &\hspace*{1.9cm} Baseline  & \hspace*{1.5cm}  Ours & \hspace*{1.0cm}Baseline   & Ours
    \end{tabularx}
\caption{Qualitative comparisons of texture between our method and the baseline method which uses a MeshCNN network to predict vertex color.}
\label{fig:color} 
\end{figure*}
\section{Comparison for Texture Generation}
Our model produces a high-resolution texture for each mesh, by reconstructing a dense texture map from the sparse UV color map with the 2D-prior network. In this section, we compare our method to a texture generation baseline method which directly predicts color for each point in the MeshCNN \cite{hanocka2019meshcnn} framework. 
On a high-level, this method uses 3D-prior only to recover the texture compared to our method that uses both 2D and 3D prior.
The network architecture of the baseline is the same as our 3D-Prior network. Instead of producing the displacement of vertices, the MeshCNN is fed with the predicted mesh shape to build the graph and trained to predict the RGB color of each vertex. The feature
on each graph node is random initialized and the loss function is \begin{equation}
    L_{color}=\sum_{\hat{p}} \|C_{\hat{p}}-C_q\|,
\end{equation}
where $q$ is the closest vertex in the input point cloud for each $\hat{p}$, and $C_{\hat{p}}$ and $C_q$ are the color estimated
for $\hat{p}$ and observed on $q$. Note that the $\hat{p}$ is randomly sampled from the predicted mesh as described in Section 3.2 of the paper, and $C_{\hat{p}}$ is calculated by linearly interpolating the predicted color of three vertexes on the corresponding triangle mesh face. Finally, the color on mesh faces is calculated with vertex color via linear interpolation. 

The qualitative results are shown in Fig. \ref{fig:color}. The quality of this baseline results is highly restricted by the vertex number of the predicted triangle mesh. Moreover, it's observed that the baseline method tends to be blurry and lose high frequency information. In contrast, our method always produces texture of high visual quality. 

\begin{figure}
\centering
 \includegraphics[width=0.4\textwidth]{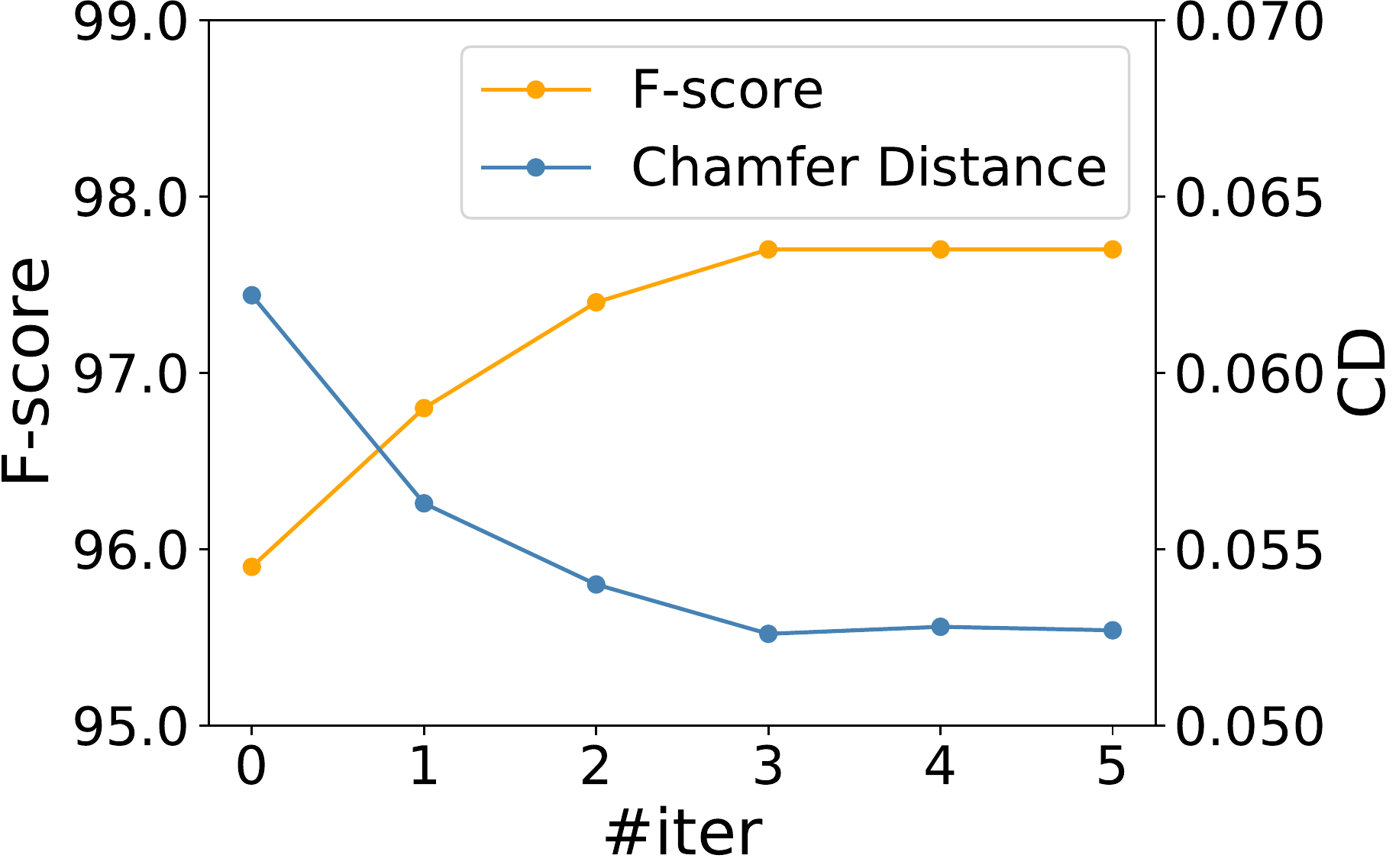}
\vspace{-4mm}
\caption{\label{fig:iter} Performance w.r.t iteration.}
\vspace{-7mm}
\end{figure}

\begin{figure*}[htb]
\centering \includegraphics[trim=0 70 0 140, clip, width=0.9\linewidth]{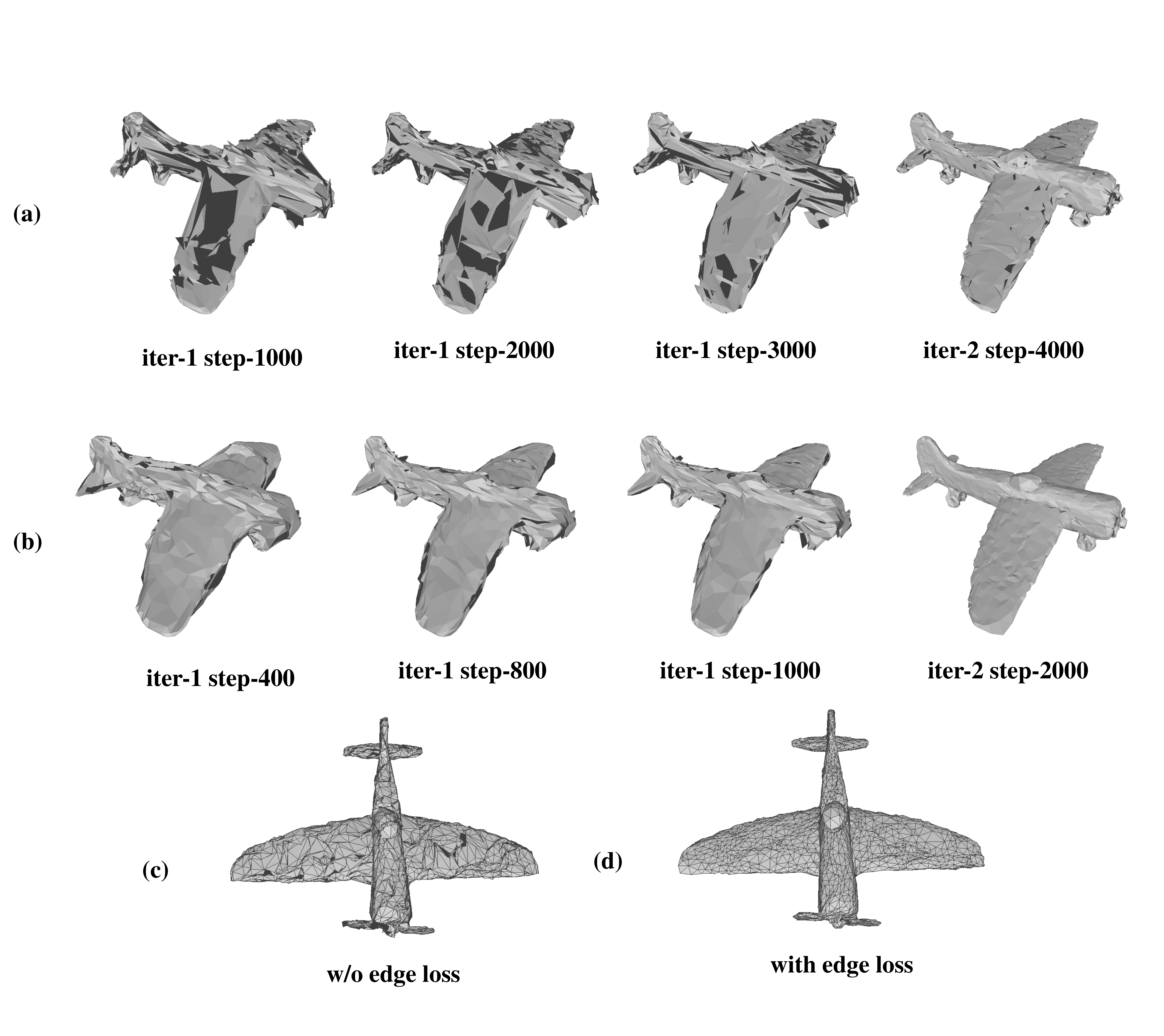}
\caption{Results (a) without edge loss (b) with edge loss under different iteration steps and the final reconstruction mesh (c)without edge loss, (d)with edge loss.}
\label{fig:edge} 
\end{figure*}

\begin{figure*}[htb]
\centering \includegraphics[width=0.88\linewidth]{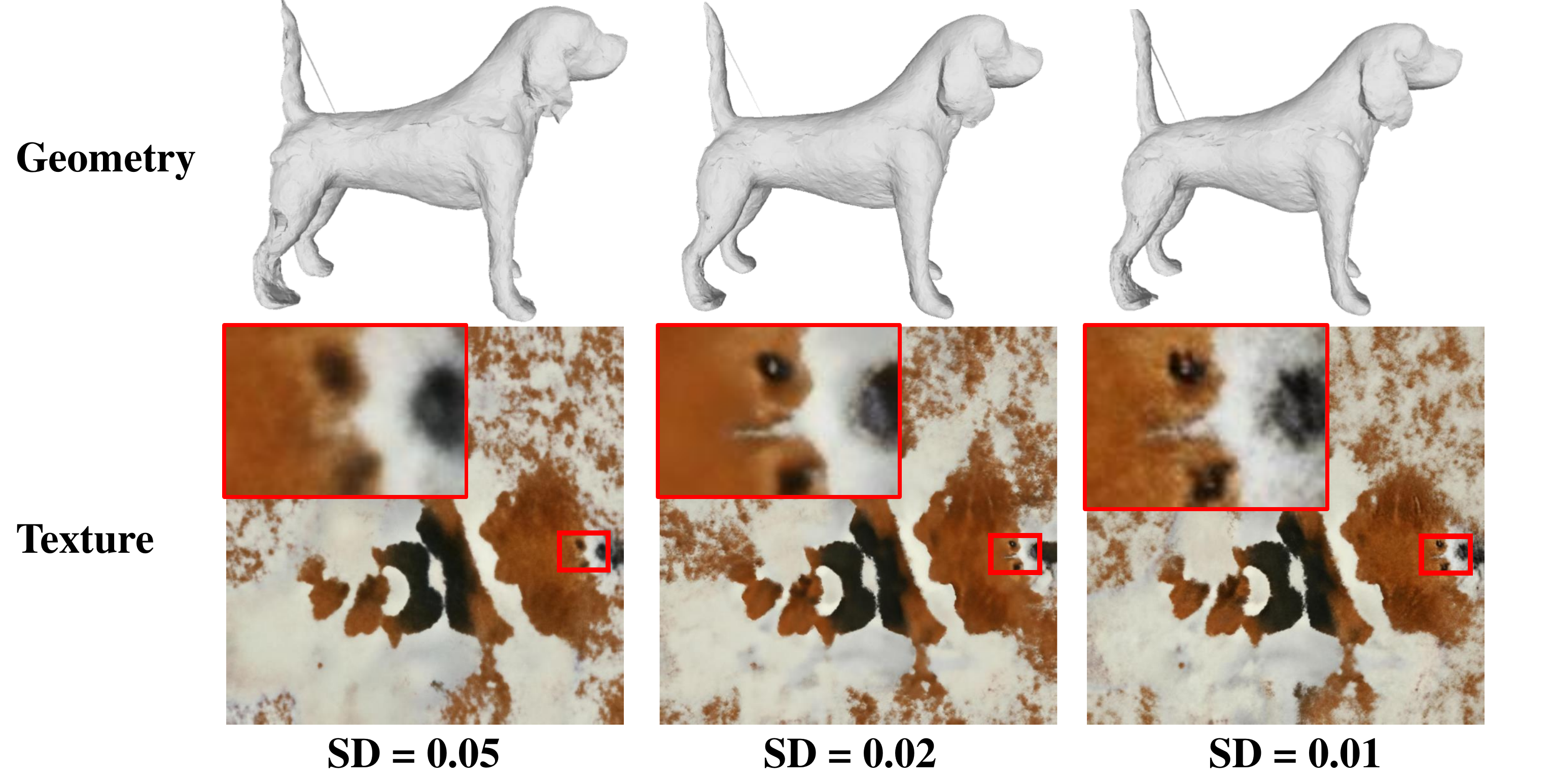}
\caption{Geometry and texture outputs with perturbation $\epsilon$ sampled from different standard deviations (SD).}
\label{fig:perbuts} 
\end{figure*}

\section{Ablation Studies}
In this section, we provide more ablation studies of our method.

\xk{\subsection{Performance w.r.t Iteration}

As shown in Fig.~\ref{fig:iter}, we report F-score and Chamfer Distance of each iteration to measure the convergence. Typically the numbers stabilize in 3 iterations.

} 
\subsection{Effect of Edge Loss in 3D-Prior Network}
As mentioned in Section 3.1, we add a loss term to constrain the edge length for the MeshCNN, which speeds up the converging speed.
In Figure \ref{fig:edge} (a) (b), we show the output of 3D-prior network w/o or w edge loss under different iteration steps 
and the final reconstruction mesh w/o or w edge loss in  Figure \ref{fig:edge} (c) (d).
Under the same iteration, the output mesh from 3D-prior network optimized with edge loss apparently exhibit better geometry, e.g. less holes and folded faces, smoother surface, compared to the case without edge loss .
Overall, to achieve similar mesh quality we get in 1000 steps using the edge loss, the network without the edge loss would need at least 4000 steps.

The edge loss can be also calculated very efficiently with known topology thus intrigues negligible computational cost to the optimization.
Overall, we can speed up the convergence of MeshCNN for 3-4 times.

\subsection{Effect of Gaussian Permutation in 2D-Prior Network}

As illustrated in Section 3.2.2 of our main submission, our 2D-Prior network takes as input a random noise feature map $z+\epsilon$, and the $\epsilon$ serves as a Gaussian permutation in each training step to prevent the network from overfitting. In this section, we show some qualitative
comparison results in Figure \ref{fig:perbuts} with $\epsilon$ sampled from different standard deviations. 

As shown in Figure \ref{fig:perbuts}, a larger permutation makes the result smoother, but also lose high frequency information. On the contrary, the network with smaller permutation tends to be overfitting. In practice, we choose 0.02 as the standard deviation of $\epsilon$ for both XYZ map and texture map generation. 

\end{document}